%% file: main.tex
\documentclass[10pt]{article} 
\usepackage[preprint]{tmlr}

\input{math_commands.tex}

\usepackage{hyperref}
\usepackage{url}
\usepackage{marvosym}

\usepackage{booktabs}       
\usepackage{amsfonts}       
\usepackage{nicefrac}       
\usepackage{microtype}      
\usepackage{xcolor}         
\usepackage[textsize=scriptsize]{todonotes} 

\usepackage{amssymb}
\usepackage{mathtools}
\usepackage{textcomp}
\usepackage{multirow}
\usepackage[table]{xcolor}
\usepackage{tcolorbox}
\usepackage{booktabs}
\usepackage{graphicx}
\usepackage{wrapfig}
\usepackage{caption}
\usepackage{colortbl}
\usepackage{subcaption}
\usepackage{bm}
\usepackage{tcolorbox}
\usepackage{float}
\definecolor{Gray}{gray}{0.93}
\usepackage{pifont}
\usepackage{fontawesome5}

\newcommand{\baseline}{TwSG}

\title{Think with Structured Grounding: Perceptual Reinforcement Learning for Chart and Visual-Tabular Understanding}

\newcommand{\blfootnote}[1]{%
  \begingroup
  \renewcommand{\thefootnote}{}%
  \footnotetext{#1}%
  \addtocounter{footnote}{-1}%
  \endgroup
}
\author{
\normalfont
\makebox[\textwidth][c]{
\begin{tabular}{c}
\textbf{Changjiang Jiang}$^{1}$ \quad
\textbf{Qiannian Zhao}$^{1}$ \quad
\textbf{Lei Xin}$^{1}$ \quad
\textbf{Jinxiang Xie}$^{2}$ \quad
\textbf{Preslav Nakov}$^{1}$ \quad
\textbf{Zhuohan Xie}$^{1}$
\\[0.5em]
\small
$^{1}$Mohamed bin Zayed University of Artificial Intelligence
\qquad
$^{2}$Nanjing University
\\[0.3em]
\end{tabular}
}
}

\def\month{MM}  
\def\year{YYYY} 
\def\openreview{\url{https://openreview.net/forum?id=XXXX}} 

\begin{document}

\maketitle

\blfootnote{
Changjiang Jiang, Qiannian Zhao, Lei Xin: This work was completed during an internship at MBZUAI. \\
\Letter~\{thejiangcj,zhuohan.xie18\}@gmail.com
}

\input{sec/0_abs}
\input{sec/1_int}
\input{sec/2_rel}
\input{sec/3_met}
\input{sec/4_exp}
\input{sec/5_con}




\bibliography{main}
\bibliographystyle{tmlr}

\appendix
\input{sec/app}

\end{document}

%% file: math_commands.tex
\usepackage{amsmath,amsfonts,bm}

\def\eqref#1{equation~\ref{#1}}

\def\1{\bm{1}}

\DeclareMathAlphabet{\mathsfit}{\encodingdefault}{\sfdefault}{m}{sl}
\SetMathAlphabet{\mathsfit}{bold}{\encodingdefault}{\sfdefault}{bx}{n}



%% file: sec/0_abs.tex
\begin{abstract}
Multimodal Large Language Models (MLLMs) capable of ``thinking with images'' often rely on external tools for fine-grained perception. However, this reliance introduces significant inference latency and fails to effectively resolve the spatial-structural gap—a fundamental challenge in text-dense and structurally relational visuals (e.g., charts and visual tables) where strict relative spatial arrangements bind textual elements. Without external tools, standard MLLMs struggle with such fine-grained visual reasoning tasks. To address these issues, we propose Think with Structured Grounding (TwSG), a novel fine-grained image perception framework designed to internalize complex images's tool-use capabilities within the model. TwSG distills the benefits of multi-step reasoning and micro-cropping into a single efficient forward pass during inference. Specifically, we use an MLLM to identify key regions guided by ground-truth answers, and then prompt a teacher model to generate high-quality visual question-answering (VQA) data. These fine-grained, region-based supervisory signals are subsequently distilled back into the full-image representation. Our training pipeline consists of two stages: (1) a cold-start supervised fine-tuning (SFT) phase using multi-turn data with focused area descriptions to foster complex reasoning and error recovery; and (2) a reinforcement fine-tuning (RFT) phase driven by a novel process reward mechanism, TL-GRPO, which encourages strategic reasoning. Extensive experiments across various MLLM architectures demonstrate that TwSG reduces inference latency while substantially improving accuracy and robustness, endowing models with native fine-grained region description and flexible reasoning capabilities.
\end{abstract}

%% file: sec/1_int.tex
\section{Introduction}


Charts and visual-tabular data, as the most prevalent mediums for structured data visualization, are ubiquitous in real-world core applications~\citep{chartqapro}. Recently, Multimodal Large Language Models (MLLMs) have demonstrated remarkable capabilities in open-ended visual question answering and understanding~\citep{chartr1}. However, existing methods for Chart Question Answering (ChartQA)~\citep{chartqa,ChartMoE} often overlook the text-dense and structurally relational nature of these formats. Specifically, charts typically feature abundant textual elements bound by strict relative spatial arrangements, creating what we term the \textit{Spatial-Structural Gap}. This challenge is equally prevalent in visual-tabular QA tasks~\citep{visualtableqa}. Previous studies have introduced ``thinking with images'' paradigms~\citep{vacot}, empowering models to invoke external tools such as dynamic cropping. While these approaches mitigate the initial perception gap, they inadvertently introduce a prohibitive efficiency bottleneck: frequent tool calling and repetitive patch inputs lead to a visual token explosion, resulting in severe inference latency and error accumulation during multi-turn interactions~\citep{zwz}. Even when an MLLM perceives fine-grained details, natural language remains inherently too ambiguous to precisely navigate dense spatial layouts. Consequently, during the Chain-of-Thought (CoT) process, the model frequently fails to consistently anchor its linguistic reasoning to specific visual coordinates, leading to logical collapse and ``spatial hallucinations'' (e.g., misattributing a value to the incorrect row or bar).

To address these issues, we propose \baseline, a structured visual data distillation and reasoning framework. Our core idea is to shift the ``interleaved visual reasoning process,'' which originally relied on external tools during the inference stage, into parameter internalization during the training stage. Specifically, when invoking external high-precision teacher models and image scaling tools, we directly distill the complex multi-turn image tool invocation generated by the teacher model into a single forward pass of the student model. To further unlock the MLLM's CvTR ability potential, we introduce a Reinforcement Learning (RL) phase following the cold-start. However, we observe that previous methods typically compute importance sampling ratios at either the token or sequence level, failing to account for the extreme variance in length across different functional tags. Specifically, the dense text information within \texttt{think} tags often results in much longer sequences compared to the concise logic in \texttt{answer} tags. This disparity leads to a \textbf{length-induced gradient bias}, where the optimization process is dominated by perceptual segments, potentially diluting the gradient contribution of critical reasoning steps. In addition, we propose TL-GRPO, a specialized reinforcement learning algorithm for \baseline's reasoning pattern. We introduce \textbf{Tag-level Importance Sampling (TL-IS)}, which normalizes importance weights independently within each tag to ensure the optimization is invariant to length disparities. To further stabilize the training, we design \textbf{Clipped Group Sampling (CGS)} to filter statistical outliers in advantage estimation. This synergy, combined with an interleaved verifiable reward mechanism, provides fine-grained guidance for the complex reasoning process, encouraging the spontaneous emergence of more strategic multi-step reasoning capabilities.

Our main contributions are summarized as follows:

\begin{itemize}
    \item We systematically identify the issues of current MLLMs in CvTR tasks. Specifically, we reveal the inherent flaws associated with the recent ``think with images'' paradigm that heavily relies on external tool invocations such as prohibitive inference latency, token explosion, and error accumulation.
    \item We propose \textbf{\baseline}, a novel paradigm that internalizes tool capabilities via data distillation and process reward-driven reinforcement learning. Extensive experiments demonstrate that our approach significantly enhances reasoning accuracy and robustness while drastically reducing inference latency compared to state-of-the-art (SOTA) methods in the domain.
    \item We introduce \textbf{TL-GRPO}, a specialized reinforcement learning algorithm tailored for structured thinking trajectories. By integrating \textit{Tag-level Importance Sampling}, \textit{Clipped Group Sampling}, and an \textit{Interleaved Verifiable Reward} mechanism, TL-GRPO effectively mitigates length-induced gradients bias and stabilizes the optimization of complex, multi-turn reasoning paths.
\end{itemize}

%% file: sec/2_rel.tex
\section{Related Work}

\input{img/img_data_pipeline}

\subsection{CvTR}
The field of Chart and Visual-Tabular Reasoning (CvTR) aims to empower MLLMs with the ability to interpret and reason over data-intensive visual structures. Early research in this domain primarily focused on single-modality table understanding~\citep{jiang2025tabdsr} or text-based TableQA~\citep{wang2024chain}. However, the emergence of benchmarks like ChartQA~\citep{chartqa} and its more sophisticated extension, ChartQAPro~\citep{chartqapro}, has shifted the focus toward interpreting complex visual marks (e.g., bars, sectors) and performing multi-step arithmetic. Parallelly, TableVQA-Bench~\citep{tableqabench} transitioned traditional text-based tabular datasets into the visual-tabular domain, demanding models to possess both high-fidelity text Recognition and spatial logical reasoning capabilities.

General-purpose MLLMs, such as Qwen3-VL~\citep{qwen3vl} and MiniCPM-V~\citep{minicpm}, have demonstrated remarkable zero-shot performance on basic chart tasks. To further push the boundaries, domain-specific models like Chart-R1~\citep{chartr1} and Chart-RVR~\citep{chartrvr} have been introduced, achieving state-of-the-art (SOTA) results through specialized fine-tuning. Additionally, CodeVision~\citep{codevision} explored a tool-augmented approach by converting image operations and cropping into executable code within a Chain-of-Thought (CoT) framework. Despite these advances, existing methods often treat chart and tabular reasoning as isolated tasks. As observed in our experiments, models like Chart-RVR exhibit a performance trade-off, where optimization for charts leads to a degradation in visual-tabular understanding. Even MoE-based approaches like ChartMoE~\citep{ChartMoE}, while effective for multi type chart scenarios, do not explicitly resolve this cross-format tension. Our work, \baseline, addresses this gap by seeking a unified reasoning perspective for both data formats, which are frequently co-present in critical domains like finance.

\subsection{Interleaved Multimodel Chain-of-Thought} 
The concept of ``Thinking with Images'' has recently gained traction as a means to alleviate the limitations of MLLMs in perceiving fine-grained visual details. This paradigm typically involves the model autonomously calling image-cropping or enhancement tools to zoom into local regions, thereby improving grounding accuracy. For instance, in text perceptional tasks, VACoT~\citep{vacot} proposed integrating image data augmentation tools into the reasoning process to resolve ambiguities in dense text recognition. Recent advancements have further diversified these visual interaction strategies: DeepEyes~\citep{deepeyes} introduced the foundational approach of invoking specialized tools for adaptive image cropping, while DeepEyesV2~\citep{deepeyesv2} evolved this into a more flexible framework by utilizing code execution to perform precise cropping and information retrieval. Furthermore, ThyME~\citep{thyme} emphasized the necessity of temporal coherence in visual reasoning, employing multi-round cropping tool invocations to iteratively refine the model's perception.

However, a wide spectrum of prior efforts—including direct QA and forgery localization methods (e.g., IML~\citep{qu2023towards}, IML2~\citep{qu2024towards}, RTM~\citep{qu2025revisiting}, Omni-IML~\citep{qu2026omni}, DS-Net~\citep{qu2026detect}, Mesorch~\citep{zhu2025mesoscopic}, Imdl-benco~\citep{ma2024imdl}, Forensichub~\citep{du2026forensichub}, RIML~\citep{zhu2026revisiting}, and Venus-DeFakerOne~\citep{DeFakerOne}) as well as broader multimodal modeling frameworks (e.g., UniMoMo~\citep{xin2026unimomo}, Beyond Human Annotation~\citep{ying2026beyond}, IMG~\citep{lin2025audio}, DualCPT~\citep{xin2026dualcpt}, Multi-Omics~\citep{xin2024artificial}, TRS~\citep{kong2025token} and Hytrec~\citep{xin2026hytrec})—primarily investigate MLLM performance within standard, direct-QA or fixed-input paradigms. While these approaches have achieved notable empirical success on closed benchmarks, they heavily rely on monolithic representations without explicit intermediate visual verification. As a result, their generalization ability proves inherently limited when transferred to structured, multi-hop reasoning tasks that demand fine-grained visual-symbolic alignment.
 
While iMCoT has shown promise in general visual grounding~\citep{qi2026patchcue}, its application to the CvTR domain remains underexplored. CvTR tasks are inherently ``text-dense'' and ``computation-intensive'', requiring both precise value extraction from small visual elements and high-level logical synthesis. Existing iMCoT frameworks have not yet been optimized for the structured, graph-like nature of charts and tables. In this paper, we bridge this gap by introducing a structured graph-based ``Thinking with Images'' strategy, enabling \baseline~to perform more robust and interpretable reasoning on complex visual data structures. The complete training parameters setup can be see in Appendix~\ref{sec:train}.

%% file: img/img_data_pipeline.tex
\begin{figure}[th]
  \centering
  \includegraphics[width=\linewidth]{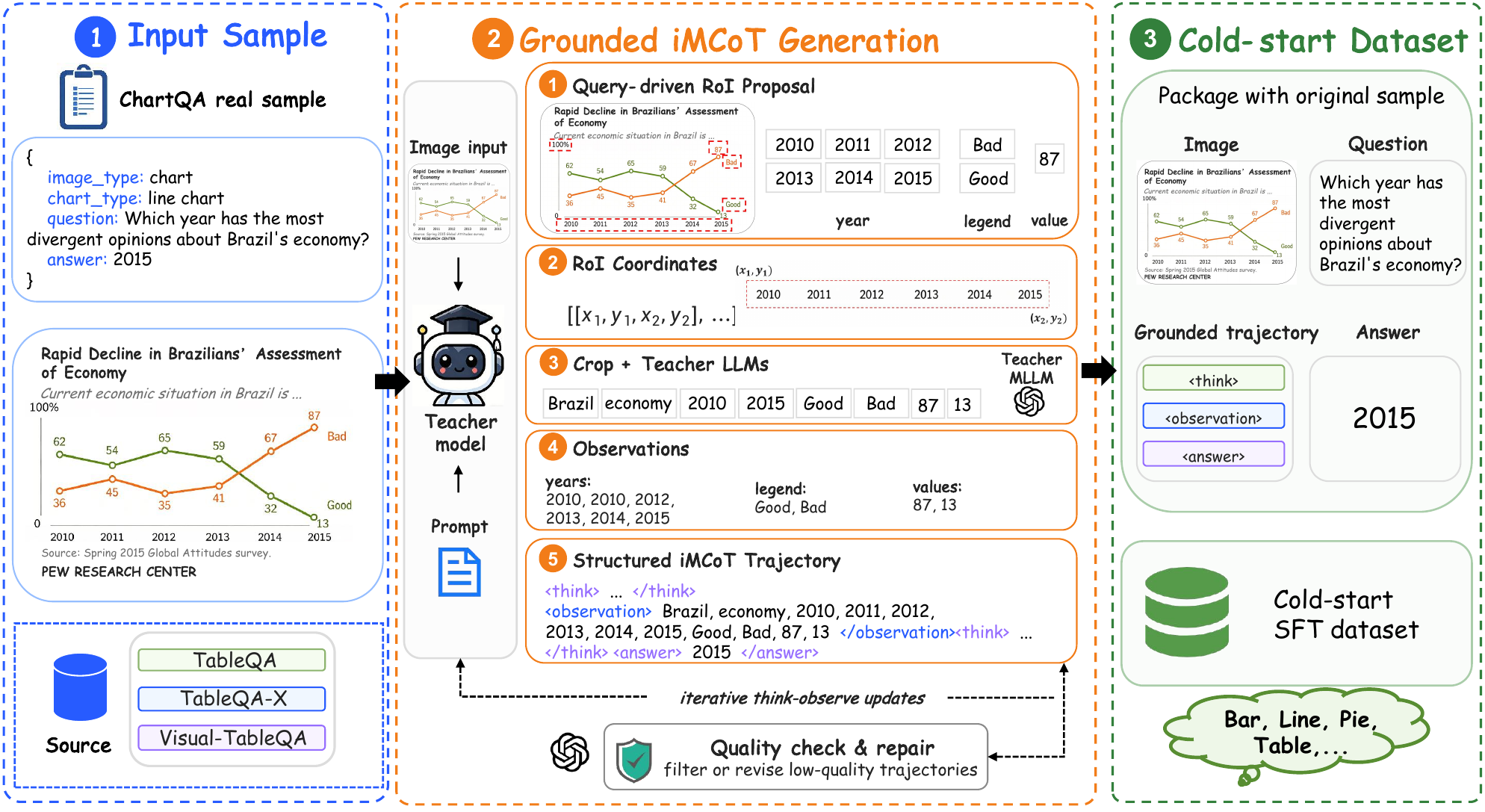}
  \caption{Pipeline for cold-start SFT data generation.}
  \label{fig:data_pipeline}
\end{figure}

%% file: sec/3_met.tex
\section{Methodology}


In this section, we introduce \textbf{Think with Structured Grounding (TwSG)}, a novel framework designed to bolster the CvTR reasoning of MLLMs on complex document data through an internalized, self-consistent cognitive process.

\subsection{Cold-start SFT Data Construction}

The core philosophy of TwSG is to leverage a powerful teacher model to perform fine-grained, region-based perception and reasoning, and then condense this process into a student model that can perform ``one-pass'' reasoning without external tools. As shown in Figure~\ref{fig:data_pipeline}, the TwSG pipeline consists of four main stages: (1) \textbf{Query-Driven Region Proposal:} Based on the given question and the global input image, a robust teacher model is employed to predict and return the most probable regions of interest (RoIs) that contain the target information; (2) \textbf{Sub-image Cropping and Teacher MLLMs Integration:} The retained RoIs are cropped into high-resolution sub-images to preserve fine-grained visual details. A teacher model is then invoked to extract highly accurate textual data from these specific crops; (3) \textbf{Structured iMCoT Trajectory Generation:} The teacher model outputs a detailed reasoning path along with the final answer. We summarize and format this output into a structured, iMCoT sequence. Crucially, to handle complex queries, this process accommodates multiple iterations of thinking and observing before reaching a conclusion. This encapsulates the internal cognitive reasoning and external perceptual steps using explicit tags, resulting in a multi-turn trajectory patterned as: \texttt{<think>} \textit{[reasoning step 1]} \texttt{</think>} \texttt{<observation>} \textit{[visual cue 1]} \texttt{</observation>} \texttt{<think>} \textit{[reasoning step 2]} \texttt{</think>} \texttt{<observation>} \textit{[visual cue 2]} \texttt{</observation>} \dots \texttt{<answer>} \textit{[final answer]} \texttt{</answer>}; (4) \textbf{Hallucination-Aware Refined Distillation:} To minimize hallucination in distilled trajectories, we utilize an independent teacher model as a Reward Model to audit the reasoning process. Instead, we instruct the teacher model to generate a corrective reasoning path. If an inconsistency is detected, rather than simply discarding the sample, we instruct the teacher model to generate a corrective reasoning path. Specifically, each ``observation'' is formatted as a structured JSON object, comprising bbox coordinates and the associated text fields, ensuring a precise mapping between visual grounding and textual evidence. Finally, we collect 12,674 Cold-Start SFT samples, denoted as \baseline-12K. Detailed data distributions and construction prompts are provided in Appendix~\ref{sec:cold_start}.



\input{img/img_2_tlgrpo}
Unlike traditional CoT which provides a single reasoning block~\citep{zwz}, we adopt an interleaved \textit{Think-Observation-Answer} pattern~\citep{icot}. Specifically, for a query $Q$, the teacher generates a trajectory $\tau = \{t_1, o_1, t_2, o_2, \dots, a\}$, where: (1) $t_i$ (\textit{Think}): The reasoning step focusing on a specific part of the structured graph $G$; (2) $o_i$ (\textit{Intermediate Evidence}): The extracted region-based perceptional result or sub-calculation result corresponding to $t_i$; (3) $a$ (\textit{Final Answer}): The final answer to the question.
This format ensures that the model grounds each reasoning step in explicit visual evidence before proceeding to the next logical deduction.

\subsection{TL-GRPO}

\paragraph{Tag-level Importance Sampling.}

As shown in Figure~\ref{fig:tl_grpo}, standard reinforcement learning frameworks, such as GRPO~\citep{grpo} and GSPO~\citep{gspo}, typically compute importance sampling ratios at either the granular token level or the sequence level. Although several recent works, such as EA-RLVR~\citep{earlvr}, Se-GUI~\citep{cite_segui}, FakeVLM-R1~\citep{cite_fakevlmr1}, EGPO~\citep{zhao2026know}, Veritas~\citep{veritas}, and Veritas++~\citep{veritasplus}, have modified multi-turn sampling schemes or reward functions, they still do not apply importance sampling from the visual Chain-of-Thought (CoT) perspective. Other approaches either adopt coarse-grained multi-turn rollout strategies, such as Fake-HR1~\citep{fakehr1}, or reformulate the overall GRPO objective along the visual CoT sequence, such as Ivy-Fake~\citep{jiang2025ivy}. Crucially, none of these methods account for the significant variance in sequence length across different functional tags. In contrast, our proposed sampling paradigm bridges the spatial-structural gap and substantially improves sampling efficiency and accuracy.

Specifically, during our cold-start phase, the model is trained to internalize dense region-based perceptual information while maintaining concise logical deductions, resulting in a highly non-uniform token distribution. Standard sampling techniques tend to be biased toward these longer perceptual segments, thereby diluting the gradient contributions of critical yet concise reasoning steps. To overcome this limitation, we propose \textbf{Tag-level Importance Sampling (TL-IS)}. By segmenting the trajectory according to our structured tripartite format, TL-IS normalizes the importance weight independently within each functional tag. This ensures that the optimization process remains invariant to length disparities between cognitive reasoning and perceptual observation, enabling more balanced and stable policy updates.

Following the above motivation, we represent a structured trajectory as a concatenation of tag-level sequences:
$\tau = \{t_1, o_1, t_2, o_2, \dots, a\}$, where each segment corresponds to a structured reasoning stage.
We denote each tag segment as $\tau_i^{(j)}$ with length $|\tau_i^{(j)}|$.

We first compute \textbf{the sequence importance sampling ratio} within each tag segment:
\begin{equation}
w_{i}^{(j)}
=
\left[
\frac{\pi_\theta (\tau_i^{(j)} \mid x)}
{\pi_{\theta_{\mathrm{old}}} (\tau_i^{(j)} \mid x)}
\right]^{\frac{1}{|\tau_i^{(j)}|}}
=
\exp\left(
\frac{1}{|\tau_i^{(j)}|}
\sum_{t \in \tau_i^{(j)}}
\log
\frac{
\pi_\theta (y_{i,t} \mid x, y_{i,<t})
}{
\pi_{\theta_{\mathrm{old}}} (y_{i,t} \mid x, y_{i,<t})
}
\right).
\end{equation}

We then aggregate across all tag segments to obtain the final tag-level importance weight:
\begin{equation}
w_i^{\mathrm{TL\text{-}GRPO}}
=
\frac{1}{S_i}
\sum_{j=1}^{S_i}
w_{i}^{(j)},
\end{equation}
where $S_i$ is the number of valid tag segments in $\tau_i$.


\paragraph{Reward function.} 
To guide the model's reasoning process and ensure output quality, we design a multi-dimensional reward function $R(y)$ consisting of three components: format integrity, answer accuracy, and region-based perceptional grounding. Formally, the total reward is defined as:
\begin{equation}
    R(y) = \lambda_1 r_{\text{fmt}} + \lambda_2 r_{\text{ans}} + \lambda_3 r_{\text{verify}}
\end{equation}
where: (1) \textbf{Format Reward ($r_{\text{fmt}}$):} We define $r_{\text{fmt}}$ as an indicator function, where $\mathcal{F}$ is the set of sequences that strictly adhere to the template $\langle\text{think}\rangle\langle\text{observation}\rangle\langle\text{answer}\rangle$ without any additional or unauthorized tags; (2) \textbf{Answer Reward ($r_{\text{ans}}$):} This term evaluates the semantic similarity between the predicted answer $y_{\text{ans}}$ and the ground truth $\hat{y}$ using ANLS, i.e., $r_{\text{ans}} = \text{ANLS}(y_{\text{ans}}, \hat{y})$; (3) \textbf{Verify Reward ($r_{\text{verify}}$):} To ensure the model remains grounded in the visual evidence, we leverage an expert MLLM to evaluate the correctness of the intermediate observation $y_{\text{observation}}$ given the image $I$. Specifically, follow by OpenAI's practice~\citep{openai2024customjudge}, we formulate observation verification as a multiple-choice evaluation task with five options, A-E.
For each intermediate observation, the judge MLLM assigns a positive score only when selecting Option A or B, which indicates that the observation is sufficiently accurate and visually grounded. 
All other options are treated as incorrect or hallucinated observations and receive a score of $0$. 
Formally, given a trajectory containing $N$ observation fields, the verification reward is computed as
\begin{equation}
r_{\mathrm{verify}} = \frac{1}{N}\sum_{i=1}^{N} \mathbb{I}\left(a_i \in \{A, B\}\right),
\end{equation}
where $a_i$ denotes the judge's selected option for the $i$-th observation. 
This normalization constrains $r_{\mathrm{verify}}$ to $[0,1]$ regardless of the number of generated observations.

This sparse yet reliable reward penalizes ungrounded observations and suppresses hallucination propagation, while providing process-level supervision for generating high-quality reasoning traces. 
During training, we use Qwen3-VL-72B-Instruct~\citep{qwen3vl} as the judge MLLM to provide verification signals for process-level reward estimation. 
The detailed judgment prompt is provided in Appendix~\ref{sec:app_verify_reward}.

\paragraph{Clipped Group Sampling.}


To further stabilize the reinforcement learning process, especially for tasks with high variance such as OCR and mathematical reasoning, we introduce \textbf{Clipped Group Sampling (CGS)}. Inspired by the dynamic sampling in DAPO~\citep{dapo} and advantage filtering in CPPO~\citep{cppo}, CGS refines the advantage distribution by trimming statistical outliers within each group. Specifically, for a group of $n$ sampled trajectories, we first compute their raw relative advantages $\hat{A}_i$. We then exclude the $k$ trajectories with the highest and lowest advantage scores (typically $k=1$), retaining the central $n-2k$ trajectories for the policy update. This ensures that the advantage distribution remains centered and symmetric, preventing the model from over-fitting to anomalous lucky successes or being distracted by singular catastrophic failures.
For the data used in TL-GRPO, following DeepSeek-R1~\citep{deepseekr1}, we applied rejection sampling using the post-SFT checkpoint across the original SFT dataset and the training sets of ChartQA~\citep{chartqa}, ChartQA-X~\citep{ChartQA-X}, and Visual-TableQA~\citep{visualtableqa}. Specifically, we generated responses four times per sample, discarding instances that were correctly answered in all attempts and retaining only those with at least one incorrect response. Ultimately, this yielded a final dataset of 64,334 entries for RFT. By synergizing these two mechanisms, our framework achieves a more robust credit assignment, effectively bridging the gap between fine-grained image perception and high-level logical deduction.

%% file: img/img_2_tlgrpo.tex
\begin{figure}[th]
  \centering
  \includegraphics[width=\linewidth]{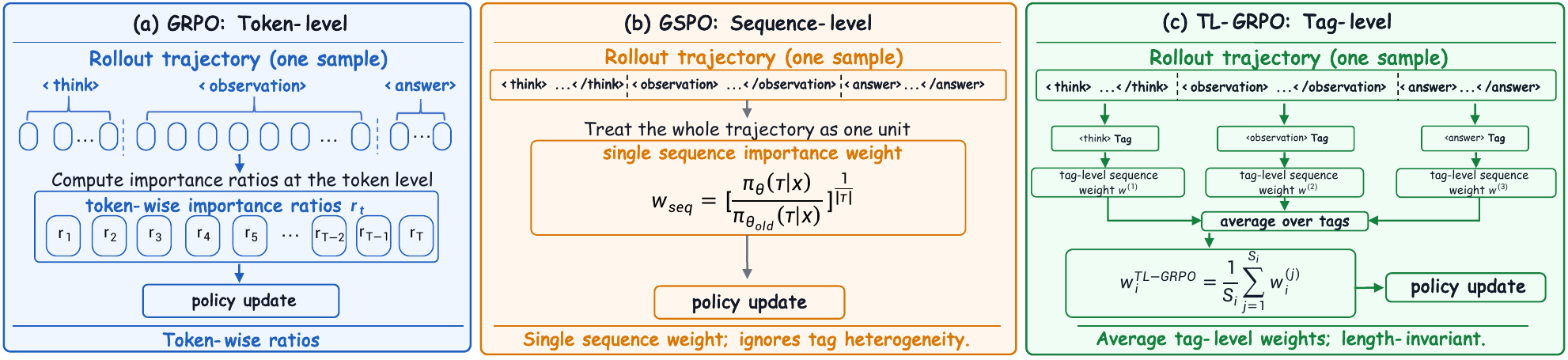}
  \caption{Comparison between Tag-level Importance Sampling and token-level or sequence-level importance sampling.}
  \label{fig:tl_grpo}
\end{figure}

%% file: sec/4_exp.tex
\section{Experiment}



\subsection{Experiment Setup}

\paragraph{Benchmarks.} We evaluate the performance of \baseline~across diverse CvTR tasks, we conduct experiments on three benchmarks: (1) TableVQA-Bench \cite{tableqabench}: This benchmark focuses on open-domain visual tabular reasoning over complex tabular data; (2) ChartQA \cite{chartqa}: A large-scale benchmark designed for question answering about charts that involves both visual and logical reasoning. It combines human-written and machine-generated questions, necessitating multi-step arithmetic operations and the ability to link visual marks (e.g., bars, lines) to their corresponding data values; (3) ChartQAPro \cite{chartqapro}: As a more challenging extension of ChartQA, ChartQAPro incorporates a higher degree of visual and topical diversity, including real-world infographics and complex dashboards. Additionally, we compare our performance with closed-source models on CharXiv-R \cite{charxiv} in Appendix~\ref{sec:ap_sota}.

\paragraph{Baselines.} We evaluate MLLMs on these representative categories: (1) General MLLMs, including Qwen3-VL-Instruct \cite{qwen3vl}, Qwen2.5-VL \cite{bai2025qwen25vltechnicalreport}, MiniCPM-V-4.5 \cite{minicpm}, and LLaVA-OneVision-1.5 \cite{an2025llavaonevision15fullyopenframework}; (2) Reasoning MLLMs, including M2-Reasoning \cite{m2reasoning}, VL-Rethinker \cite{vlrethinker}, and Visionary-R1 \cite{visionaryr1}; (3) Chart and Visual-Tabular (CvTR)-domain MLLMs, including Chart-R1 \cite{chartr1} and Chart-RVR \cite{chartrvr}. \baseline's backbone is Qwen3-VL-8B.



\input{tab/main_tab}

\subsection{Main Result}


As shown in Table~\ref{tab:id}, we compare MLLMs at comparable model
scales on a unified suite of chart and visual-tabular reasoning (CvTR)
benchmarks. \baseline-8B achieves the best overall performance, with an
average accuracy of \textbf{73.70\%}, outperforming the strongest prior model in this comparison, Chart-R1, by 68.12\%. Moreover,
\baseline-8B obtains the best performance on 10 out of the 11 evaluated
subsets, demonstrating consistently strong generalization across both
chart and visual-tabular reasoning tasks.

Strong performance on standard chart benchmarks does not necessarily translate into robust performance across more challenging chart and visual-tabular reasoning tasks. General-purpose MLLMs often achieve strong accuracy on ChartQA, yet their performance does not consistently transfer to the more reasoning-intensive ChartQAPro and TableVQA-Bench subsets. In contrast, \baseline~exhibits more balanced performance across these benchmarks. Notably, even \baseline-4B achieves an average accuracy of \textbf{68.16\%}, slightly surpassing the 7B Chart-R1 model (68.12\%) despite using a smaller model scale. This further demonstrates the effectiveness of unified chart and visual-tabular reasoning.

\input{img/img_3_inference-speed}

For example, compared with its base model Qwen2.5-VL-3B, Chart-RVR-Hard improves ChartQA-H from 74.24\% to 80.08\%, while its FinTabNetQA accuracy decreases from 77.76\% to 70.16\%. These results suggest that improvements from chart-specific specialization do not necessarily transfer to structured visual-tabular data, motivating a unified treatment of the two modalities.

Figure~\ref{fig:infer} compares the inference throughput of different
reasoning paradigms. Tool-augmented approaches, particularly the
Think-with-Images series, incur substantial inference overhead due to
external tool invocation and repeated perception--reasoning interactions.
In contrast, our approach internalizes bounding-box reasoning and spatial
grounding into the model's native reasoning process, avoiding external
tool calls during inference. \baseline~therefore maintains approximately
1.75--2.00 samples per second and achieves higher throughput than the
Qwen3-VL base model, while substantially improving CvTR performance.
Together with the accuracy results in Table~\ref{tab:id}, these results
demonstrate a favorable accuracy--efficiency trade-off through unified,
tool-free end-to-end reasoning.

\subsection{Ablation Study}


\input{tab/main_close_source}

We conduct comprehensive ablation studies on ChartQAPro and TableVQA-Bench to validate the key design choices of our framework, with detailed results summarized in Table~\ref{tab:all_ablation}.

\paragraph{Comparison of TL-GRPO and Baseline.} 
Both TL-IS and CGS consistently yield performance gains over the standard GRPO baseline. When combined, TL-GRPO achieves superior performance, outperforming GRPO by +6.20\% on ChartQAPro (from 49.40\% to 55.60\%) and +8.99\% on TableVQA-Bench (from 77.80\% to 86.79\%). This gain stems from a critical domain characteristic in visual reasoning: reasoning trajectories ($\langle\texttt{think}\rangle$) are disproportionately long relative to the final prediction ($\langle\texttt{answer}\rangle$). Under standard token-level weighting, long reasoning steps dilute the gradient signal of compact answers, which tag-level sequence importance sampling effectively mitigates.

\paragraph{Effect of Cold Start SFT.} 
The SFT checkpoint without RL (\textit{w/o $R$}) achieves 48.22\% on ChartQAPro and 78.53\% on TableVQA-Bench. While cold start establishes fundamental instruction-following and question-answering capabilities, its generalization on complex visual-tabular reasoning remains limited without reinforcement learning.

\paragraph{Reward Function Decomposition.} 
Evaluating individual reward terms reveals that all components are essential. Supervising solely with outcome correctness ($R_{\text{ans}}$) leads to severe performance degradation across both benchmarks (dropping to 47.30\% and 60.22\%), indicating optimization instability under sparse and noisy answer-only rewards. Incorporating formatting constraints ($R_{\text{fmt}}$) substantially restores and improves accuracy to 52.30\% and 82.91\%. Further adding visual verification ($R_{\text{verify}}$) achieves the best performance (55.60\% / 86.79\%), yielding additional gains of +3.30\% and +3.88\% respectively, which confirms the importance of fine-grained verification for structured reasoning.

\paragraph{Sensitivity of Clipping Hyperparameter $K$.}
Setting $K=0$ corresponds to training with TL-IS alone without advantage-based trajectory clipping (achieving 52.34\% and 83.33\%). The optimal trade-off is achieved at $K=1$ (55.60\% / 86.79\%). As $K$ increases to 2 and 3, performance consistently degrades, because filtering out too many extreme-advantage samples excessively suppresses gradient variance and slows policy optimization.

\paragraph{Contribution of Reasoning Tags.}
Dissecting intermediate reasoning tokens shows distinct dependencies across tasks. Removing the $\langle\texttt{think}\rangle$ tag incurs the largest performance drop on ChartQAPro (-7.37\%), while removing the $\langle\texttt{observation}\rangle$ tag causes the sharpest drop on TableVQA-Bench (-6.84\%), verifying that explicit visual grounding is indispensable for structured tabular data. In contrast, removing $\langle\texttt{answer}\rangle$ results in a modest yet consistent degradation (-1.28\% / -1.29\%), demonstrating the utility of explicit answer boundary tokens in stabilizing output decoding.



%% file: tab/main_tab.tex
\begin{table}[tb]
\caption{
Performance comparison on chart and visual-tabular reasoning benchmarks.
Accuracy (\%) is reported for the machine-generated (M) and human-generated
(H) subsets of ChartQA, five question types of ChartQAPro, and four data
sources of TableVQA-Bench. Avg.\ denotes the macro-average over all 11 subsets.
}
  \label{tab:id}
  \centering
  \resizebox{0.975\textwidth}{!}{
  \begin{tabular}{lccc|cccccccccc}
    \toprule
        \multirow{3}{*}{\textbf{Method}} & \multirow{3}{*}{\textbf{Size}} & \multicolumn{7}{c}{\textbf{Chart}} & \multicolumn{4}{c}{\textbf{Visual Tabular}} & \multirow{3}{*}{Avg.} \\
    
        \cmidrule(lr){3-9} \cmidrule(lr){10-13} 

        &  & \multicolumn{2}{c}{ChartQA} & \multicolumn{5}{c}{ChartQAPro} & \multicolumn{4}{c}{TableVQA-Bench} &  \\

        & & M & H & Factoid & MCQ & Convers. & FactChk. & Hypoth. & VWTQ & VWTQ-Syn & VTabFact & FinTabNetQA & \\
        \midrule
        \multicolumn{14}{c}{\textbf{General MLLMs}} \\
        & 4B & \underline{95.44} & 78.80 & 37.34 & 43.46 & 31.48 & 52.46 & 43.86 & 60.11 & 65.55 & 83.20 & 77.58 & 60.84 \\
        \multirow{-2}{*}{LLaVA-OneVision-1.5} & 8B & 94.64 & 78.88 & 38.88 & 35.05 & 34.72 & 53.28 & 46.38 & 62.20 & 71.33 & 84.00 & 76.82 & 61.47 \\
        & 3B & 94.24 & 74.24 & 29.24 & 38.32 & 33.86 & 49.59 & 38.91 & 56.27 & 65.59 & 75.60 & 77.76 & 57.60 \\
        \multirow{-2}{*}{Qwen2.5-VL} & 7B & 94.88 & 80.56 & 38.88 & 49.53 & 34.66 & 56.56 & 37.73 & 61.67 & 70.61 & 80.00 & 72.88 & 61.63\\
        & 4B & 94.24 & 73.28 & 29.24 & 38.32 & 33.86 & 49.59 & 38.91 & 56.17 & 61.77 & 76.80 & 73.19 & 56.85 \\
        \multirow{-2}{*}{Qwen3-VL} & 8B & 94.72 & 76.40 & 39.53 & 55.14 & 36.81 & 60.25 & 39.83 & 59.56 & 63.29 & 80.80 & 77.05 & 62.13\\
        MINICPM-V-4.5 & 8.7B & 81.60 & 67.44 & \textbf{47.39} & 48.13 & 41.16 & 56.56 & 39.54 & 71.48 & 76.85 & 89.60 & \underline{78.80} & 63.50 \\
        \midrule
        \multicolumn{14}{c}{\textbf{Reasoning MLLMs}} \\
        Visionary-R1 & 3B & 92.40 & 69.84 & 27.41 & 31.31 & 31.57 & 36.48 & 36.58 & 50.63 & 62.10 & 74.80 & 76.33 & 53.59 \\
        M2-Reasoning & 7B & 88.80 & 80.32 & 37.68 & 54.67 & \underline{41.93} & 53.28 & \underline{55.90} & 72.51 & 78.65 & \underline{90.00} & 76.43 & 66.38 \\
        VL-Rethinker & 7B & 83.12 & 72.72 & 43.36 & \underline{60.75} & 41.27 & 60.66 & 53.78 & 68.41 & 70.79 & 86.00 & 72.68 & 64.87 \\
        \midrule
        \multicolumn{14}{c}{\textbf{CvTR-domain MLLMs}} \\
        Chart-RVR & 3B & 93.76 & 75.76 & 30.07 & 51.40 & 32.58 & 49.59 & 37.05 & 54.37 & 63.91 & 81.20 & 68.10 & 57.98\\
        Chart-RVR-Hard & 3B & 92.88 & 80.08 & 30.88 & 54.67 & 36.32 & 50.00 & 40.32 & 54.97 & 62.89 & 83.60 & 70.16 & 59.71 \\
        Chart-R1 & 7B & 95.20 & \underline{88.72} & 40.63 & 52.80 & 40.62 & \underline{63.93} & 51.71 & 71.39 & 77.98 & 87.60 & 78.74 & 68.12 \\
         & 4B 
        & 95.36 & 86.48 
        & 37.88 & 58.92 & 40.73 & 61.02 & 50.91 
        & \underline{72.83} & \underline{79.66} & 88.40 & 77.52 
        & \underline{68.16} \\
        
        \multirow{-2}{*}{\baseline} & 8B 
        & \textbf{96.12} & \textbf{89.44} 
        & \underline{45.92} & \textbf{62.35} & \textbf{44.80} & \textbf{66.18} & \textbf{58.77} 
        & \textbf{79.93} & \textbf{88.72} & \textbf{95.60} & \textbf{82.92} 
        & \textbf{73.70} \\
  \bottomrule
  \end{tabular}
}
\end{table}

%% file: img/img_3_inference-speed.tex
\begin{figure}[h]
    \centering
    \includegraphics[width=0.6\linewidth]{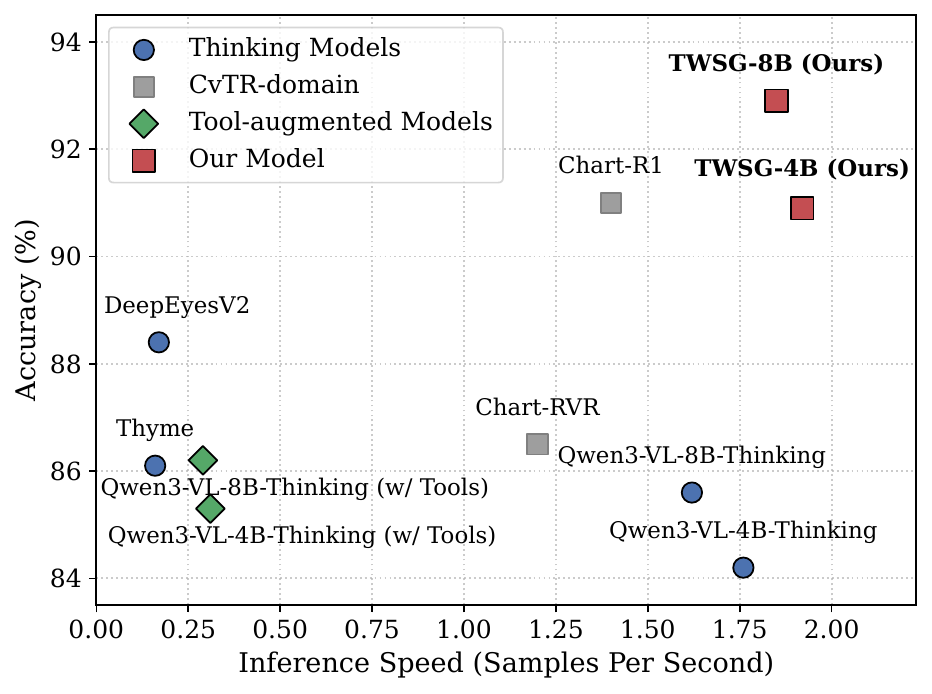}
    \caption{Samples per second.}
    \label{fig:infer}
\end{figure}

%% file: tab/main_close_source.tex


\begin{table}[h]
    \centering
\caption{
Ablation studies of TL-GRPO on ChartQAPro and TableVQA-Bench.
\textit{w/o $R$} denotes the SFT checkpoint before reinforcement learning.
Best results are shown in bold.
}
    \label{tab:all_ablation}

\resizebox{0.75\linewidth}{!}{
    \begin{tabular}{@{}lcccc@{}}
    \toprule

\multicolumn{5}{c}{\textit{Effect of TL-GRPO}} \\
\cmidrule(lr){1-5}
Benchmark
    & GRPO
    & GRPO + TL-IS
    & GRPO + CGS
    & TL-GRPO (Full) \\
\midrule
ChartQAPro
    & 49.40
    & 52.34
    & 53.65
    & \textbf{55.60} \\
TableVQA-Bench
    & 77.80
    & 83.33
    & 82.53
    & \textbf{86.79} \\

    \midrule

\multicolumn{5}{c}{\textit{Reward Component Ablation}} \\
\cmidrule(lr){1-5}
Benchmark
    & w/o $R$
    & $R_{\mathrm{ans}}$
    & $R_{\mathrm{ans}}+R_{\mathrm{fmt}}$
    & $R_{\mathrm{ans}}+R_{\mathrm{fmt}}+R_{\mathrm{verify}}$ \\
\midrule
ChartQAPro
    & 48.22
    & 47.30
    & 52.30
    & \textbf{55.60} \\
TableVQA-Bench
    & 78.53
    & 60.22
    & 82.91
    & \textbf{86.79} \\

    \midrule

    \multicolumn{5}{c}{\textit{Reasoning Component Removal}} \\
    \cmidrule(lr){1-5}
    Benchmark
        & $\langle$\texttt{think}$\rangle$
        & $\langle$\texttt{observation}$\rangle$
        & $\langle$\texttt{answer}$\rangle$
        & Full \\
    \midrule
    ChartQAPro
        & 48.23
        & 51.72
        & 54.32
        & \textbf{55.60} \\
    TableVQA-Bench
        & 80.92
        & 79.95
        & 85.50
        & \textbf{86.79} \\

    \midrule

    \multicolumn{5}{c}{\textit{CGS Hyperparameter $K$}} \\
    \cmidrule(lr){1-5}
    Benchmark
        & $K=0$
        & $K=1$
        & $K=2$
        & $K=3$ \\
    \midrule
    ChartQAPro
        & 52.34
        & \textbf{55.60}
        & 53.66
        & 50.02 \\
    TableVQA-Bench
        & 83.33
        & \textbf{86.79}
        & 82.12
        & 79.88 \\

    \bottomrule
    \end{tabular}
    }
\end{table}

%% file: sec/5_con.tex
\section{Conclusion}
In this paper, we identify that relying on the ``think with images'' paradigm for perceptual understanding in highly structured, text-intensive visual contexts introduces prohibitive inference latency and exacerbates the Spatial-Structural Gap in CvTR tasks. To overcome these bottlenecks, we propose \baseline, a novel structured data distillation framework tailored specifically for charts and tables. This framework significantly enhances the fine-grained OCR reasoning and perceptual capabilities of MLLMs. Furthermore, by introducing TL-GRPO and integrating a two-stage training paradigm that combines cold-start supervised fine-tuning with RL, we systematically align and bolster the model's performance in complex CvTR tasks. Extensive ablation studies and inference latency evaluations empirically demonstrate the effectiveness and efficiency of our proposed method. Ultimately, our work provides a highly efficient, tool-free pathway for advancing MLLMs in structurally complex visual reasoning.

\subsection*{Broader Impact Statement}
Due to computational resource constraints, our framework was exclusively trained and evaluated on models with 8B parameters or fewer. Additionally, our approach inherently relies on the accuracy of the raw Cold start data. Consequently, the initial context fed into the large model cannot be guaranteed to be entirely error-free. Nevertheless, our empirical results demonstrate that our proposed method can still substantially enhance the CvTR capabilities of MLLMs despite this dependency.

Furthermore, we observe that computational reasoning challenges within chart understanding remain unaddressed, as pure CoT prompting falls short in handling complex arithmetic calculations. In contrast, TabDSR~\citep{jiang2025tabdsr} demonstrates the superiority of Program-of-Thought (PoT) in enhancing computational capabilities. Inspired by CodeVision~\citep{codevision}, we plan to explore internalized, code-grounded image operations and programmatic visual reasoning in future work.

Our work aims to improve structured visual reasoning for charts and visual tables, which can benefit data analysis, document understanding, and accessibility-oriented applications. We do not develop technologies for weapons, biometric identification, surveillance, or direct decision-making in high-stakes domains. Therefore, we do not anticipate direct safety risks such as physical harm or increased weapon lethality.

%% file: sec/app.tex
\section{License}

All datasets and models used in our experiments are available for research purposes. We declare that this study is conducted solely for academic research and has no commercial purpose.

\section{Experimental Setup}
\label{sec:train}

The training of \baseline~is conducted in two distinct phases: (1) Cold-start SFT. To initialize the reasoning capability of the model, we perform SFT with a learning rate of 1e-5 and a global batch size of 32; (2) Following the cold-start SFT, we employ TL-GRPO to further refine the model's reasoning trajectory. Unlike SFT, this stage emphasizes final outcome correctness rather than dense CoT supervision. We set the reward coefficient $\beta$ to 0.0 and utilize a dual-clip epsilon strategy with $\epsilon = 3 \times 10^{-4}$ and $\epsilon_{high} = 4 \times 10^{-4}$ to stabilize importance sampling. The learning rate is decayed to $1 \times 10^{-6}$ with a warmup ratio of 0.05. For each prompt, we sample $G=8$ generations to compute the relative reward. The models are trained on a cluster of 128 NVIDIA A100 GPUs (80GB). The total computational budget for training and evaluation ranges from 50 to 120 wall-clock hours, depending on the backbone scale.

\section{Compare with close-source MLLMs}
\label{sec:ap_sota}
As shown in Table~\ref{tab:sota}, on CharXiv-R~\citep{charxiv}, \baseline-8B achieves \textbf{67.8\%}, outperforming strong proprietary models such as GPT-4.1 (56.7\%) and GPT-4.5 (55.4\%), and even surpassing large-scale models like Qwen3-VL-235B-A22B-Thinking (66.1\%) despite a significantly smaller parameter size.  Compared with recent ``thinking with images'' models, our method demonstrates superior performance without explicit test-time reasoning or tool use, indicating that our gains come from improved intrinsic multimodal reasoning rather than inference-time scaling.

\input{tab/main_chartqa_sota}

\section{Verify Reward}
\label{sec:app_verify_reward}

Evaluating the quality of open-ended intermediate reasoning steps is notoriously difficult, as traditional string-matching metrics often fail to assess semantic correctness and visual faithfulness. To address this limitation, we leverage an expert MLLM as an automated judge to compute the verify reward ($r_{\text{verify}}$). This follows the general practice of LLM/MLLM-as-a-judge evaluation, where strong foundation models are used to assess open-ended outputs beyond exact string matching.

The primary advantage of this approach is its ability to perform nuanced, visually grounded evaluation. By framing the assessment as a rigorous multiple-choice task, the expert MLLM can explicitly differentiate critical visual hallucinations from acceptable minor phrasing variations.

\input{tab/tab_app_1_reward_ab}

\input{tab/tab_sup_model-ab}

\subsection{Ablation Experiment}

\paragraph{Effect of judge models.} Table~\ref{tab:ab_reward}~(a) shows that the choice of judge MLLM for $r_{\mathrm{verify}}$ has only a marginal impact on the final performance. Replacing the default Qwen3-VL-72B-Instruct~\citep{qwen3vl} with stronger proprietary judges does not lead to consistent gains: GPT-4o obtains 54.44 on ChartQAPro and 85.32 on TableVQA-Bench, while Qwen3.6-plus~\citep{qwen36plus} reaches 56.32 and 85.90, respectively. In comparison, our default Qwen3-VL-72B-Instruct achieves 55.60 on ChartQAPro and the best result of 86.79 on TableVQA-Bench. The overall variation across different judges is small, suggesting that our reward verification is not highly sensitive to the specific judge model once the judge has sufficient multimodal reasoning capability. From a cost perspective, using Qwen3-VL-72B-Instruct is also more practical.
\paragraph{Effect of reward weights.}
Table~\ref{tab:ab_reward}~(b) analyzes the effect of different reward-weight configurations in TL-GRPO. 
When only the answer reward is used, i.e., $(\lambda_1,\lambda_2,\lambda_3)=(0.00,1.00,0.00)$, the model obtains 52.30 on ChartQAPro and 82.91 on TableVQA-Bench, indicating that direct answer supervision provides the main optimization signal. 
After introducing small weights for the format reward and verification reward, the performance improves to 54.83 and 85.74 with $(0.05,0.90,0.05)$, showing that output-format regularization and judge-based verification provide complementary guidance. 
The default setting $(0.10,0.80,0.10)$ achieves the best performance on both benchmarks, with 55.60 on ChartQAPro and 86.79 on TableVQA-Bench. This suggests that answer correctness should remain the dominant reward, while lightweight format and verification rewards are beneficial for stabilizing the reasoning process and improving final accuracy.

\paragraph{Effect of decoding hyperparameters.}
Table~\ref{tab:ab_decoding} evaluates the robustness of our method under different decoding hyperparameters. 
Across three random seeds, the performance remains stable, with only minor fluctuations on both ChartQAPro and TableVQA-Bench. 
This indicates that the reported results are not sensitive to random sampling effects. 
We also vary the sampling temperature from 0.1 to 0.9. 
The performance changes only slightly, while higher temperatures lead to a small degradation due to increased output randomness. 
These results suggest that our method is robust to decoding configurations and does not rely on carefully tuned test-time sampling.


\input{img/img_pro_judge}

\subsection{API Cost}

For the exact prompt template used to query the expert MLLM, including the detailed instructions and the complete option descriptions provided to the model, please refer to Figure~\ref{pro:verify_rewaard}.

Public API pricing shows that GPT-4o is substantially more expensive~\citep{openrouter}, commonly listed around \$2.50/M input tokens and \$10.00/M output tokens, while Qwen3.6-plus is listed around \$0.355/M input tokens and \$1.99/M output tokens. By contrast, Qwen-VL/Qwen3-VL family models are available at lower-cost API endpoints, and Qwen3-VL-72B-Instruct can also be deployed locally as an open-weight judge, further reducing large-scale training-time annotation cost. Therefore, we adopt Qwen3-VL-72B-Instruct as the default judge since it provides a better trade-off between verification quality and training cost.

\input{img/img_data_case}

\input{img/img_pro_roi}

\section{Cold-Start SFT Data Construction}
\label{sec:cold_start}
This section details the construction process of the Cold-Start SFT data, including the distillation of CvTR-domain datasets and their statistical distributions. The primary objective of our distillation process is to extract interleaved CoT from diverse data sources, enabling the MLLM to internalize this reasoning paradigm—a strategy that has been widely adopted in recent state-of-the-art methodologies~\citep{visionaryr1,deepseekr1}. Consequently, this phase focuses on constructing a high-quality offline dataset where the inputs comprise the image $I$, the question $Q$, and the ground truth label from the original training set, while the output is the corresponding elicited CoT. 

\input{img/img_area_desc}

We collect raw training data from the original training splits of TableQA, TableQA-X, and Visual-TableQA. To further enrich the visual layouts, especially for multi-chart and multi-table scenarios, we randomly synthesize composite samples consisting of two charts, two tables, or one chart and one table, while preserving the original question-answer pairs. This strategy allows us to efficiently expand the coverage of complex chart-table layouts without introducing additional annotation cost. We then perform the following filtering procedure:

Step 1: ROI Identification. The initial step involves leveraging a leading closed-source MLLM to identify all potential ROI bounding boxes. Specifically, given the full image $I$ and the question $Q$, the model determines which sub-regions are most relevant to the query and returns their absolute coordinates. For this ROI generation, we utilize GPT-4o. The detailed prompt for this task is illustrated in Figure~\ref{fig:pro_1_roi}.

Step 2: Region Description. In the second step, we crop the corresponding sub-regions based on each generated bounding box to obtain sub-images $sub\_I$. We then prompt the MLLM to generate a comprehensive description for each region. Unlike standard global description prompts, our instruction is explicitly optimized for text recognition within charts and tables (as detailed in Figure~\ref{fig:pro_2_area_description}). One area exects once time. We similarly employ GPT-5.1 for this high-fidelity content description task.

\input{img/img_pro_clean}

Step 3: CoT Refinement and Structuring. The third step refines the CoT generated in the preceding stages. We empirically observed that raw CoT often contains colloquialisms and excessive redundancy. Given the critical role of data quality, we instruct the MLLM to perform two key tasks: (1) simplify the reasoning path by preserving only essential text-related descriptions, and (2) extract textual content strictly pertinent to the question. For instance, if a query specifically targets a ``yellow region,'' any extraneous global descriptions are distilled to retain only information concerning that target area. We require the model (Qwen3.6-Plus~\citep{qwen36plus}) to produce this refined data in a structured JSON format. Finally, we encapsulate the comprehensive visual observations within \texttt{<observation>} tags, while the question-specific reasoning is enclosed within \texttt{<think>} tags, as shown in Figure~\ref{fig:pro_3_clean}.

Finally, we conduct a hallucination detection phase for each data entry to ensure faithfulness. The evaluation methodology and the corresponding verification prompt are provided in Figure~\ref{pro:verify_rewaard}. Unlike the main text, this section details the specific, step-by-step construction process.

\input{img/img_data_dist}

Figure~\ref{fig:data_case} illustrates the diversity of visualization and table types covered in our benchmark. 
The benchmark includes 45 categories, spanning standard statistical charts, distribution plots, relational diagrams, temporal visualizations, geographic maps, hierarchical structures, and complex tables. 
Specifically, it covers basic chart types such as bar, line, scatter, pie, histogram (his), heatmap, box, violin, radar, area, bubble, and 3D plots; advanced visualization forms such as error bars, error points, multi-axis charts (Axis Chart), ring charts (Ring), rose charts (Rose), treemaps, contour plots, density plots, quiver plots, funnels, stacked/grouped bars, stacked areas, waterfalls, candlesticks, Gantt charts, timelines, calendar heatmaps, Sankey diagrams, sunburst charts, maps, choropleth maps, network graphs, dendrograms, surface plots, and polar plots. 
In addition, the benchmark includes five table-oriented categories: single-column tables, cross-row tables, cross-column tables, mixed tables, and Multi chart and table layouts (Multi CvT). Following the query design of ChartQAPro~\citep{chartqapro}, we define eight question categories in our benchmark, including Mathematical Reasoning, Visual Reasoning, Conversational, Multiple-Choice, Hypothetical, Fact-Checking, Unanswerable, and Multi-Chart QA. To build a balanced benchmark, we first train a local Qwen3-32B classifier to categorize all candidate questions in the seed pool, and further train a Qwen3-VL-8B classifier to identify the chart/table type of each sample. Based on these automatic annotations, we perform category balancing and then manually filter redundant or low-quality samples. As shown in Figure~\ref{fig:data_dist}, the final benchmark contains 12,674 samples spanning 45 chart/table categories and 8 question categories. The chart and table distribution exhibits a mildly long-tailed pattern, where common categories such as bar, line, scatter, and pie occupy relatively larger portions, while rarer categories remain sufficiently represented. Meanwhile, the question distribution is relatively balanced across the eight categories, which helps ensure comprehensive evaluation over diverse reasoning skills.


\textbf{The Distillation Objective.} Given the teacher-generated interleaved trajectory $\tau$, we optimize the student model $M_\theta$ using a standard cross-entropy loss:
\begin{equation}
    \mathcal{L}_{TwSG} = - \sum_{j=1}^{|\tau|} \log P(x_j | I, Q, x_{<j}; \theta)
\end{equation}
where $x_j$ are the tokens in the interleaved sequence. By training on these trajectories, the student learns to internally simulate the ``Reason-Observation-Reason'' process.

This approach mimics the reasoning process of iteratively referencing visual charts during complex, multi-step problem-solving. Consequently, it significantly mitigates the ``coordinate hallucination'' prevalent in MLLMs~\citep{fu-etal-2025-mitigating}—instances where the predicted coordinates decouple from the reasoning context. Furthermore, by integrating an expert teacher model, our method effectively alleviates the inherent hallucinations typically observed during the model's CoT. 

\section{Qualitative Analysis}


\input{img/img_fig_case}

As shown in Figure~\ref{fig:case}, existing CvTR-domain MLLMs can identify the relevant heat-map region but often fail to establish a faithful mapping between color-coded cells and their textual values. Chart-R1 directly extracts incorrect values from non-orange cells, while Chart-RVR repeats the same error despite producing a longer reasoning trace, suggesting that extended reasoning alone does not guarantee reliable visual grounding. In contrast, TwSG first localizes the key region through the $\langle$observation$\rangle$ field and explicitly lists the orange-cell values before aggregation.

%% file: tab/main_chartqa_sota.tex
\begin{table*}[h]
\caption{Performance comparison on chart reasoning benchmarks. Left: results on ChartQA with open-source MLLMs. Right: comparison with proprietary models on the CharXiv-R subset. ``Tools'' indicates whether external tool use is enabled. \textbf{Bold} denotes the best, and \underline{underline} the second best. ChartMOE performs inference with an external Python tool.}
  \label{tab:sota}
\centering
\begin{subtable}[t]{0.52\linewidth}
  \caption{Comparison with general MLLMs on ChartQA.}
  \label{tab:chartqa_left}
  \centering
\resizebox{\linewidth}{!}{
  \begin{tabular}{lccc}
\toprule
\textbf{Model} & \textbf{Venue} & \textbf{Tools} & \textbf{ChartQA (\%)} \\
\midrule
\multicolumn{4}{c}{\textbf{General-domain MLLMs}} \\
\midrule
Qwen2.5-VL-7B~\citep{bai2025qwen25vltechnicalreport} & - & & 87.3 \\
Qwen3-VL-8B~\citep{qwen3vl} & - & & 85.6 \\
Thyme~\citep{thyme} & ICLR'26 & \checkmark & 86.1 \\
DeepEyesV2~\citep{deepeyesv2} & ICLR'26 & \checkmark & \underline{88.4} \\
VGR-7B~\citep{vgr} & ICLR'26 & \checkmark & 72.8 \\
\midrule
\multicolumn{4}{c}{\textbf{CvTR-domain MLLMs}} \\
\midrule
ChartAst-S~\citep{meng2024chartassistant} & ACL'24 & & 79.9 \\
ChartLLaMa~\citep{han2023chartllama} & - & & 69.7 \\
BigCharts-R1-7B~\citep{bigcharts} & COLM'25 & & 89.8 \\
Chart-R1~\citep{chartr1} & - & & 91.0 \\
ChartMOE*~\citep{ChartMoE} & ICLR'25 & \checkmark & 87.8 \\
\midrule
\rowcolor[gray]{0.95} \textbf{Ours (\baseline-8B)} & - &  & \textbf{92.9}\\
\bottomrule
  \end{tabular}
  }
  \end{subtable}
  \hfill
  \begin{subtable}[t]{0.47\linewidth}
  \caption{Comparison with large size MLLMs on CharXiv-R.}
  \label{tab:chartqa_r}
  \centering
\resizebox{\linewidth}{!}{
  \begin{tabular}{lcc}
\toprule
\textbf{Model} & \textbf{Provider / Venue} & \textbf{CharXiv-R (\%)} \\
\midrule
\multicolumn{3}{c}{\textbf{General-domain MLLMs}} \\
\midrule
GPT-4o & OpenAI & 47.1 \\
Qwen3-VL-8B-Thinking & Alibaba & 53.0 \\
GPT-4.5 & OpenAI & 55.4 \\
GPT-4.1 & OpenAI & 56.7 \\
Qwen3-VL-235B-A22B-Instruct & Alibaba & 62.1 \\
Qwen3-VL-32B-Instruct & Alibaba & 62.8 \\
Qwen3-VL-32B-Thinking & Alibaba & \underline{65.2} \\
Qwen3-VL-235B-A22B-Thinking & Alibaba & 66.1 \\
\midrule
\multicolumn{3}{c}{\textbf{CvTR-domain MLLMs}} \\
\midrule
BigCharts-R1-7B~\citep{bigcharts} & COLM'25 & 41.3 \\
Bespoke-MiniChart~\citep{meng2025mmeurekaexploringfrontiersmultimodal} & - & 45.4 \\
Chart-R1~\citep{chartr1} & - & 46.2 \\
\midrule
\rowcolor[gray]{0.95} \textbf{Ours (\baseline-8B)} & - & \textbf{67.8} \\
\bottomrule
  \end{tabular}
  }
  \end{subtable}
\end{table*}

%% file: tab/tab_app_1_reward_ab.tex
\begin{table*}[t]
    \scriptsize
    \centering
    \caption{Ablation studies on reward verification and reward weighting.}
    \label{tab:ab_reward}

    \begin{minipage}[t]{0.46\linewidth}
        \centering
        \resizebox{\linewidth}{!}{
            \begin{tabular}{lcc}
                \toprule
                \multicolumn{3}{c}{\textbf{(a) Judge Model}} \\
                \midrule
                Method & ChartQAPro & TableVQA-Bench \\
                \midrule
                GPT-4o & 54.44 & 85.32 \\
                Qwen3.6-plus & 56.32 & 85.90 \\
                \rowcolor{Gray}
                Qwen3-VL-72B-Instruct (Default) & 55.60 & 86.79 \\
                \bottomrule
            \end{tabular}
        }
    \end{minipage}
    \hfill
    \begin{minipage}[t]{0.46\linewidth}
        \centering
        \resizebox{\linewidth}{!}{
            \begin{tabular}{ccccc}
                \toprule
                \multicolumn{5}{c}{\textbf{(b) Reward Weights}} \\
                \midrule
                $\lambda_1$ & $\lambda_2$ & $\lambda_3$
                & ChartQAPro & TableVQA-Bench \\
                \midrule
                0.00 & 1.00 & 0.00 & 52.30 & 82.91 \\
                0.05 & 0.90 & 0.05 & 54.83 & 85.74 \\
                \rowcolor{Gray}
                0.10 & 0.80 & 0.10 & 55.60 & 86.79 \\
                \bottomrule
            \end{tabular}
        }
    \end{minipage}
\end{table*}

%% file: tab/tab_sup_model-ab.tex
\begin{table}[t]
    \centering
    \caption{Impact of decoding hyperparameters. Left: different random seeds. Right: different sampling temperatures.}
    \label{tab:ab_decoding}

    \begin{minipage}[t]{0.44\linewidth}
        \centering
        \resizebox{\linewidth}{!}{
            \begin{tabular}{ccc}
                \toprule
                \multicolumn{3}{c}{\textbf{(a) Seed}} \\
                \midrule
                Seed & ChartQAPro & TableVQA \\
                \midrule
                42   & 55.60 & 86.79 \\
                1024 & 55.47 & 86.62 \\
                8192 & 55.71 & 86.84 \\
                \bottomrule
            \end{tabular}
        }
    \end{minipage}
    \hfill
    \begin{minipage}[t]{0.44\linewidth}
        \centering
        \resizebox{\linewidth}{!}{
            \begin{tabular}{ccc}
                \toprule
                \multicolumn{3}{c}{\textbf{(b) Temperature}} \\
                \midrule
                Temp & ChartQAPro & TableVQA \\
                \midrule
                0.1 & 55.60 & 86.79 \\
                0.5 & 55.42 & 86.51 \\
                0.9 & 55.18 & 86.35 \\
                \bottomrule
            \end{tabular}
        }
    \end{minipage}
\end{table}

%% file: img/img_pro_judge.tex
\begin{figure}[h]
\begin{tcolorbox}[
  colback=white,
  colframe=black,
  coltext=black,
  boxrule=0.5mm,
  fontupper=\ttfamily\footnotesize,
  width=\linewidth
]
\textbf{System Instruction:} \\
You are an expert visual evaluator. Your task is to carefully assess the accuracy of a generated ``Observation'' based strictly on the provided image. Compare the textual description with the visual evidence and determine its level of accuracy.

\textbf{Input:} \\
- \textbf{Image:} [Sub Image] \\
- \textbf{Observation:} \texttt{[Visual Description]}

\textbf{Evaluation Options:} \\
Please select the most appropriate option from the following categories:
\begin{itemize}
    \item[\textbf{A.}] \textbf{Completely Accurate:} The observation perfectly aligns with the visual content, containing absolutely no hallucinations or factual errors.
    \item[\textbf{B.}] \textbf{Mostly Accurate:} The observation is largely correct and visually grounded, with only minor or negligible discrepancies that do not impact the overall understanding.
    \item[\textbf{C.}] \textbf{Partially Accurate:} The observation contains a mix of accurate visual descriptions and noticeable hallucinations or errors.
    \item[\textbf{D.}] \textbf{Mostly Inaccurate:} The observation is dominated by severe hallucinations, or fundamentally contradicts the primary visual evidence.
    \item[\textbf{E.}] \textbf{Completely Unrelated:} The observation completely fails to describe the image, or is entirely irrelevant to the visual content.
\end{itemize}

\textbf{Output Format:} \\
Provide your judgment by outputting \textbf{ONLY} a single uppercase letter corresponding to your choice (A, B, C, D, or E). Do not include any explanations, punctuation, or extra text.
\end{tcolorbox}
  \caption{Prompt for Verify Reward.}
  \label{pro:verify_rewaard}
\end{figure}

%% file: img/img_data_case.tex
\begin{figure}[h]
  \centering
  \includegraphics[width=\linewidth]{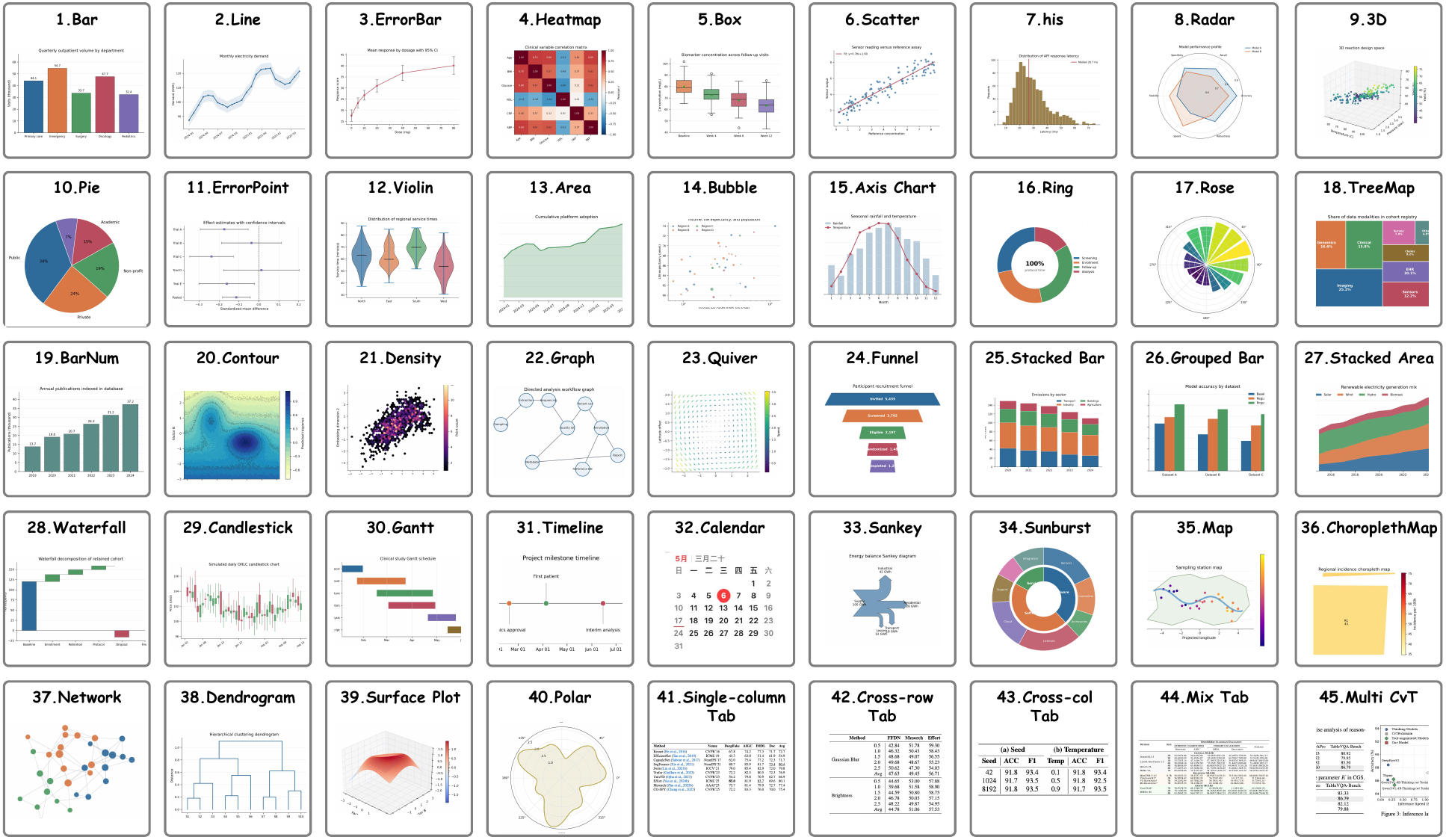}
  \caption{Illustrative examples of the 45 chart and table categories covered in our benchmark.}
  \label{fig:data_case}
\end{figure}

%% file: img/img_pro_roi.tex
\begin{figure}[h]
\centering
\begin{tcolorbox}[
  colback=white,
  colframe=black,
  coltext=black,
  boxrule=0.5mm,
  width=\linewidth
]
\footnotesize
\ttfamily
You are a visual localization assistant. Given a question and an input image, identify the region(s) of interest (ROIs) that contain the visual evidence needed to answer the question.

\vspace{0.5em}

Do not answer the question. Return only the relevant bounding box(es) and a brief description of what each region contains.

\vspace{0.5em}

Return the result as a JSON list of dictionaries only:

\begin{verbatim}
[
  {
    "bbox_2d": [x1, y1, x2, y2],
    "region_summary": "A brief description of the localized region."
  },
  ...
]
\end{verbatim}

Requirements:
\begin{enumerate}
    \item Each dictionary corresponds to one question-relevant ROI.
    \item The bounding box must tightly cover the relevant region.
    \item The coordinates must follow the format [x1, y1, x2, y2].
    \item The region summary should be brief and factual.
    \item Do not extract detailed text or numbers at this stage.
    \item Do not provide the final answer.
    \item Do not hallucinate regions that are not visible in the image.
\end{enumerate}

Question: \{question\}

Image: \{image\}
\end{tcolorbox}
\caption{Prompt for ROI extraction.}
\label{fig:pro_1_roi}
\end{figure}

%% file: img/img_area_desc.tex
\begin{figure}[h]
\centering
\begin{tcolorbox}[
  colback=white,
  colframe=black,
  coltext=black,
  boxrule=0.5mm,
  width=\linewidth
]
\footnotesize
\ttfamily

You are a visual reasoning assistant.

Given a image summary and an input image, identify the image region that contains information useful for answering the question.

Describe the given region and explicitly include the key visible text/numbers from that region.

Return JSON only:

\begin{verbatim}
{
  "bbox_2d": [x1, y1, x2, y2],
  "area_description": "Describe the given region, including its spatial context and all 
  question-relevant visible text/numbers exactly as shown in the image."
}
\end{verbatim}

Requirements:
\begin{enumerate}
    \item Include exact visible text/numbers in \texttt{area\_description}.
    \item Include labels, row/column names, cell values, axis values, or legends if relevant.
    \item List multiple relevant values explicitly.
    \item Do not hallucinate missing text.
\end{enumerate}

Input:

Image Summary: \{region\_summary\}

Image: \{image\}
\end{tcolorbox}
\caption{Prompt for region description and text extraction.}
\label{fig:pro_2_area_description}
\end{figure}

%% file: img/img_pro_clean.tex
\begin{figure}[h]
\centering
\begin{tcolorbox}[
  colback=white,
  colframe=black,
  coltext=black,
  boxrule=0.5mm,
  width=\linewidth
]
\footnotesize
\ttfamily

You are a CoT refinement assistant. Given a question, an input image, and raw textual descriptions extracted from multiple localized regions, clean and structure the reasoning information for high-quality data construction.

Your task is to simplify the raw descriptions and retain only the information that is directly useful for answering the question. Remove colloquial expressions, redundant descriptions, irrelevant global context, and any content unrelated to the question. For each localized region, output a concise question-relevant textual description.

Return JSON only:

\begin{verbatim}
{
    "cleaned_description": "A concise description of the question-relevant
    text, numbers, labels, or visual evidence in this region."
}
\end{verbatim}

Requirements:
\begin{enumerate}
    \item Preserve only essential information that is directly related to the question.
    \item Extract textual content strictly pertinent to the question.
    \item Remove redundant, colloquial, or irrelevant descriptions.
    \item If the question targets a specific region, such as a yellow region, keep only information about that target region.
    \item Do not add information that is not visible in the image or not present in the raw text.
    \item Do not answer the question.
    \item The output must be a JSON dictionary.
\end{enumerate}

Input:

Question: \{question\}

Sub Image: \{image\}

Raw region descriptions: \{raw\_region\_descriptions\}

\end{tcolorbox}
\caption{Prompt for CoT refinement and structuring.}
\label{fig:pro_3_clean}
\end{figure}

%% file: img/img_data_dist.tex
\begin{figure}[h]
  \centering
  \includegraphics[width=\linewidth]{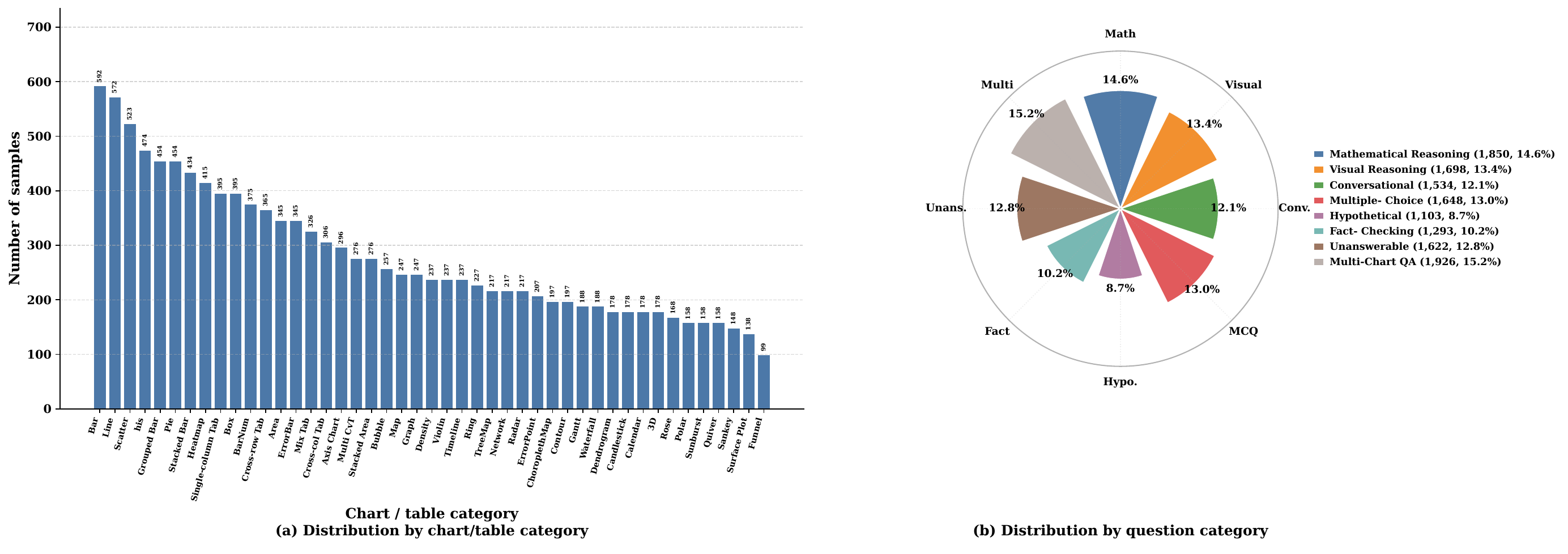}
  \caption{The Static of \baseline-12K.}
  \label{fig:data_dist}
\end{figure}

%% file: img/img_fig_case.tex
\begin{figure}[tb]
  \centering
  \includegraphics[width=\linewidth]{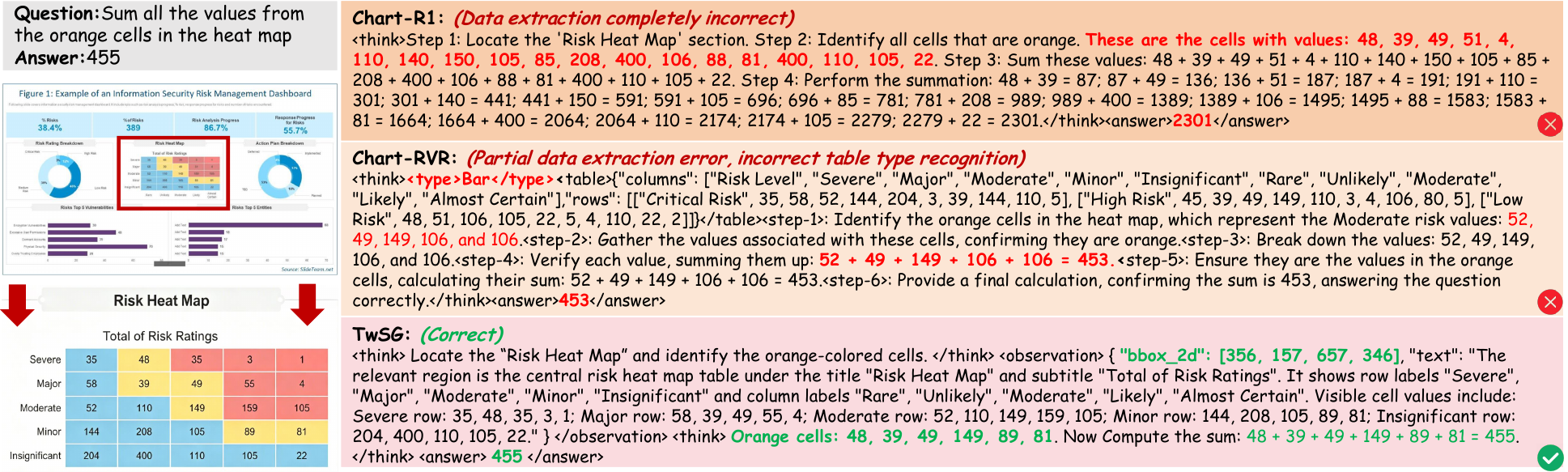}
  \caption{Reasoning comparisons between \baseline~and existing CvTR-domain MLLMs.}
  \label{fig:case}
\end{figure}

%% file: main.bib
@String(CVPR  = {IEEE Conf. Comput. Vis. Pattern Recog.})

@String(NeurIPS = {Adv. Neural Inform. Process. Syst.})

@String(ICLR  = {Int. Conf. Learn. Represent.})

@String(AAAI  = {AAAI})

@String(ICASSP=	{ICASSP})

@String(CVPR  = {CVPR})

@String(NeurIPS = {NeurIPS})

@String(ICLR  = {ICLR})

@misc{minicpm,
      title={MiniCPM-V 4.5: Cooking Efficient MLLMs via Architecture, Data, and Training Recipe}, 
      author={Tianyu Yu and Zefan Wang and Chongyi Wang and Fuwei Huang and Wenshuo Ma and Zhihui He and Tianchi Cai and Weize Chen and Yuxiang Huang and Yuanqian Zhao and Bokai Xu and Junbo Cui and Yingjing Xu and Liqing Ruan and Luoyuan Zhang and Hanyu Liu and Jingkun Tang and Hongyuan Liu and Qining Guo and Wenhao Hu and Bingxiang He and Jie Zhou and Jie Cai and Ji Qi and Zonghao Guo and Chi Chen and Guoyang Zeng and Yuxuan Li and Ganqu Cui and Ning Ding and Xu Han and Yuan Yao and Zhiyuan Liu and Maosong Sun},
      year={2025},
      eprint={2509.18154},
      archivePrefix={arXiv},
      primaryClass={cs.LG},
      url={https://arxiv.org/abs/2509.18154}, 
}

@misc{an2025llavaonevision15fullyopenframework,
      title={LLaVA-OneVision-1.5: Fully Open Framework for Democratized Multimodal Training}, 
      author={Xiang An and Yin Xie and Kaicheng Yang and Wenkang Zhang and Xiuwei Zhao and Zheng Cheng and Yirui Wang and Songcen Xu and Changrui Chen and Didi Zhu and Chunsheng Wu and Huajie Tan and Chunyuan Li and Jing Yang and Jie Yu and Xiyao Wang and Bin Qin and Yumeng Wang and Zizhen Yan and Ziyong Feng and Ziwei Liu and Bo Li and Jiankang Deng},
      year={2025},
      eprint={2509.23661},
      archivePrefix={arXiv},
      primaryClass={cs.CV},
      url={https://arxiv.org/abs/2509.23661}, 
}

@inproceedings{
visualtableqa,
title={Visual-Table{QA}: Open-Domain Benchmark for Reasoning over Table Images},
author={Boammani Aser Lompo and Marc Haraoui},
booktitle={NeurIPS 2025 Workshop on Foundations of Reasoning in Language Models},
year={2025},
url={https://openreview.net/forum?id=fvJRsGwhPf}
}

@misc{tableqabench,
      title={TableVQA-Bench: A Visual Question Answering Benchmark on Multiple Table Domains}, 
      author={Yoonsik Kim and Moonbin Yim and Ka Yeon Song},
      year={2024},
      eprint={2404.19205},
      archivePrefix={arXiv},
      primaryClass={cs.CV},
      url={https://arxiv.org/abs/2404.19205}, 
}

@inproceedings{chartqapro,
    title = "{C}hart{QAP}ro: A More Diverse and Challenging Benchmark for Chart Question Answering",
    author = "Masry, Ahmed  and
      Islam, Mohammed Saidul  and
      Ahmed, Mahir  and
      Bajaj, Aayush  and
      Kabir, Firoz  and
      Kartha, Aaryaman  and
      Laskar, Md Tahmid Rahman  and
      Rahman, Mizanur  and
      Rahman, Shadikur  and
      Shahmohammadi, Mehrad  and
      Thakkar, Megh  and
      Parvez, Md Rizwan  and
      Hoque, Enamul  and
      Joty, Shafiq",
    editor = "Che, Wanxiang  and
      Nabende, Joyce  and
      Shutova, Ekaterina  and
      Pilehvar, Mohammad Taher",
    booktitle = "Findings of the Association for Computational Linguistics: ACL 2025",
    month = jul,
    year = "2025",
    address = "Vienna, Austria",
    publisher = "Association for Computational Linguistics",
    url = "https://aclanthology.org/2025.findings-acl.978/",
    doi = "10.18653/v1/2025.findings-acl.978",
    pages = "19123--19151",
    ISBN = "979-8-89176-256-5"}

@misc{chartrvr,
      title={Chart-RVR: Reinforcement Learning with Verifiable Rewards for Explainable Chart Reasoning}, 
      author={Sanchit Sinha and Oana Frunza and Kashif Rasul and Yuriy Nevmyvaka and Aidong Zhang},
      year={2025},
      eprint={2510.10973},
      archivePrefix={arXiv},
      primaryClass={cs.CV},
      url={https://arxiv.org/abs/2510.10973}, 
}

@misc{vlrethinker,
      title={VL-Rethinker: Incentivizing Self-Reflection of Vision-Language Models with Reinforcement Learning}, 
      author={Haozhe Wang and Chao Qu and Zuming Huang and Wei Chu and Fangzhen Lin and Wenhu Chen},
      year={2025},
      eprint={2504.08837},
      archivePrefix={arXiv},
      primaryClass={cs.LG},
      url={https://arxiv.org/abs/2504.08837}, 
}

@misc{visionaryr1,
      title={Visionary-R1: Mitigating Shortcuts in Visual Reasoning with Reinforcement Learning}, 
      author={Jiaer Xia and Yuhang Zang and Peng Gao and Sharon Li and Kaiyang Zhou},
      year={2025},
      eprint={2505.14677},
      archivePrefix={arXiv},
      primaryClass={cs.CV},
      url={https://arxiv.org/abs/2505.14677}, 
}

@misc{ChartQA-X,
      title={ChartQA-X: Generating Explanations for Visual Chart Reasoning}, 
      author={Shamanthak Hegde and Pooyan Fazli and Hasti Seifi},
      year={2025},
      eprint={2504.13275},
      archivePrefix={arXiv},
      primaryClass={cs.CV},
      url={https://arxiv.org/abs/2504.13275}, 
}

@inproceedings{chartqa,
    title = "{C}hart{QA}: A Benchmark for Question Answering about Charts with Visual and Logical Reasoning",
    author = "Masry, Ahmed  and
      Long, Do Xuan  and
      Tan, Jia Qing  and
      Joty, Shafiq  and
      Hoque, Enamul",
    editor = "Muresan, Smaranda  and
      Nakov, Preslav  and
      Villavicencio, Aline",
    booktitle = "Findings of the Association for Computational Linguistics: ACL 2022",
    month = may,
    year = "2022",
    address = "Dublin, Ireland",
    publisher = "Association for Computational Linguistics",
    url = "https://aclanthology.org/2022.findings-acl.177/",
    doi = "10.18653/v1/2022.findings-acl.177",
    pages = "2263--2279"}

@misc{qwen3vl,
      title={Qwen3-VL Technical Report}, 
      author={Shuai Bai and Yuxuan Cai and Ruizhe Chen and Keqin Chen and Xionghui Chen and Zesen Cheng and Lianghao Deng and Wei Ding and Chang Gao and Chunjiang Ge and Wenbin Ge and Zhifang Guo and Qidong Huang and Jie Huang and Fei Huang and Binyuan Hui and Shutong Jiang and Zhaohai Li and Mingsheng Li and Mei Li and Kaixin Li and Zicheng Lin and Junyang Lin and Xuejing Liu and Jiawei Liu and Chenglong Liu and Yang Liu and Dayiheng Liu and Shixuan Liu and Dunjie Lu and Ruilin Luo and Chenxu Lv and Rui Men and Lingchen Meng and Xuancheng Ren and Xingzhang Ren and Sibo Song and Yuchong Sun and Jun Tang and Jianhong Tu and Jianqiang Wan and Peng Wang and Pengfei Wang and Qiuyue Wang and Yuxuan Wang and Tianbao Xie and Yiheng Xu and Haiyang Xu and Jin Xu and Zhibo Yang and Mingkun Yang and Jianxin Yang and An Yang and Bowen Yu and Fei Zhang and Hang Zhang and Xi Zhang and Bo Zheng and Humen Zhong and Jingren Zhou and Fan Zhou and Jing Zhou and Yuanzhi Zhu and Ke Zhu},
      year={2025},
      eprint={2511.21631},
      archivePrefix={arXiv},
      primaryClass={cs.CV},
      url={https://arxiv.org/abs/2511.21631}, 
}

@misc{bai2025qwen25vltechnicalreport,
      title={Qwen2.5-VL Technical Report}, 
      author={Shuai Bai and Keqin Chen and Xuejing Liu and Jialin Wang and Wenbin Ge and Sibo Song and Kai Dang and Peng Wang and Shijie Wang and Jun Tang and Humen Zhong and Yuanzhi Zhu and Mingkun Yang and Zhaohai Li and Jianqiang Wan and Pengfei Wang and Wei Ding and Zheren Fu and Yiheng Xu and Jiabo Ye and Xi Zhang and Tianbao Xie and Zesen Cheng and Hang Zhang and Zhibo Yang and Haiyang Xu and Junyang Lin},
      year={2025},
      eprint={2502.13923},
      archivePrefix={arXiv},
      primaryClass={cs.CV},
      url={https://arxiv.org/abs/2502.13923}, 
}

@misc{chartr1,
      title={Chart-R1: Chain-of-Thought Supervision and Reinforcement for Advanced Chart Reasoner}, 
      author={Lei Chen and Xuanle Zhao and Zhixiong Zeng and Jing Huang and Yufeng Zhong and Lin Ma},
      year={2025},
      eprint={2507.15509},
      archivePrefix={arXiv},
      primaryClass={cs.AI},
      url={https://arxiv.org/abs/2507.15509}, 
}

@misc{m2reasoning,
      title={M2-Reasoning: Empowering MLLMs with Unified General and Spatial Reasoning}, 
      author={Inclusion AI and : and Fudong Wang and Jiajia Liu and Jingdong Chen and Jun Zhou and Kaixiang Ji and Lixiang Ru and Qingpei Guo and Ruobing Zheng and Tianqi Li and Yi Yuan and Yifan Mao and Yuting Xiao and Ziping Ma},
      year={2025},
      eprint={2507.08306},
      archivePrefix={arXiv},
      primaryClass={cs.AI},
      url={https://arxiv.org/abs/2507.08306}, 
}

@article{han2023chartllama,
  title={Chartllama: A multimodal llm for chart understanding and generation},
  author={Han, Yucheng and Zhang, Chi and Chen, Xin and Yang, Xu and Wang, Zhibin and Yu, Gang and Fu, Bin and Zhang, Hanwang},
  journal={arXiv preprint arXiv:2311.16483},
  year={2023}
}

@inproceedings{meng2024chartassistant,
  title={ChartAssistant: A universal chart multimodal language model via chart-to-table pre-training and multitask instruction tuning},
  author={Meng, Fanqing and Shao, Wenqi and Lu, Quanfeng and Gao, Peng and Zhang, Kaipeng and Qiao, Yu and Luo, Ping},
  booktitle={Findings of the Association for Computational Linguistics: ACL 2024},
  pages={7775--7803},
  year={2024}
}

@misc{gspo,
      title={Group Sequence Policy Optimization}, 
      author={Chujie Zheng and Shixuan Liu and Mingze Li and Xiong-Hui Chen and Bowen Yu and Chang Gao and Kai Dang and Yuqiong Liu and Rui Men and An Yang and Jingren Zhou and Junyang Lin},
      year={2025},
      eprint={2507.18071},
      archivePrefix={arXiv},
      primaryClass={cs.LG},
      url={https://arxiv.org/abs/2507.18071}, 
}

@inproceedings{deepeyesv2,
author = {Jack Hong and Chenxiao Zhao and ChengLin Zhu and Weiheng Lu and Guohai Xu and Xing Yu},
title = {DeepEyesV2: Toward Agentic Multimodal Model},
booktitle = ICLR,
year = 2026
}

@inproceedings{deepeyes,
author = {Ziwei Zheng and Michael Yang and Jack Hong and Chenxiao Zhao and Guohai Xu and Le Yang and Chao Shen and Xing Yu},
title = {DeepEyes: Incentivizing "Thinking with Images" via Reinforcement Learning},
booktitle = ICLR,
year = 2026
}

@misc{thyme,
      title={Thyme: Think Beyond Images}, 
      author={Yi-Fan Zhang and Xingyu Lu and Shukang Yin and Chaoyou Fu and Wei Chen and Xiao Hu and Bin Wen and Kaiyu Jiang and Changyi Liu and Tianke Zhang and Haonan Fan and Kaibing Chen and Jiankang Chen and Haojie Ding and Kaiyu Tang and Zhang Zhang and Liang Wang and Fan Yang and Tingting Gao and Guorui Zhou},
      year={2025},
      eprint={2508.11630},
      archivePrefix={arXiv},
      primaryClass={cs.CV},
      url={https://arxiv.org/abs/2508.11630}, 
}

@misc{vacot,
      title={VACoT: Rethinking Visual Data Augmentation with VLMs}, 
      author={Zhengzhuo Xu and Chong Sun and SiNan Du and Chen Li and Jing Lyu and Chun Yuan},
      year={2025},
      eprint={2512.02361},
      archivePrefix={arXiv},
      primaryClass={cs.CV},
      url={https://arxiv.org/abs/2512.02361}, 
}

@article{wang2024chain,
  title={Chain-of-Table: Evolving Tables in the Reasoning Chain for Table Understanding},
  author={Wang, Zilong and Zhang, Hao and Li, Chun-Liang and Eisenschlos, Julian Martin and Perot, Vincent and Wang, Zifeng and Miculicich, Lesly and Fujii, Yasuhisa and Shang, Jingbo and Lee, Chen-Yu and Pfister, Tomas},
  journal={ICLR},
  year={2024}
}

@inproceedings{
vgr,
title={{VGR}: Visual Grounded Reasoning},
author={Jiacong Wang and Zijian Kang and Haochen Wang and LiangXiao and Ya Wang and Jiawen Li and Bohong Wu and Ran Jiao and Haiyong Jiang and ChaoFeng and Jun Xiao},
booktitle={The Fourteenth International Conference on Learning Representations},
year={2026},
url={https://openreview.net/forum?id=kDhAiaGzrn}
}

@inproceedings{
bigcharts,
title={BigCharts-R1: Enhanced Chart Reasoning with Visual Reinforcement Finetuning},
author={Ahmed Masry and Abhay Puri and Masoud Hashemi and Juan A. Rodriguez and Megh Thakkar and Khyati Mahajan and Vikas Yadav and Sathwik Tejaswi Madhusudhan and Alexandre Pich{\'e} and Dzmitry Bahdanau and Christopher Pal and David Vazquez and Enamul Hoque and Perouz Taslakian and Sai Rajeswar and Spandana Gella},
booktitle={Second Conference on Language Modeling},
year={2025},
url={https://openreview.net/forum?id=19fydz1QnW}
}

@inproceedings{jiang2025tabdsr,
  title={TABDSR: Decompose, Sanitize, and Reason for Complex Numerical Reasoning in Tabular Data},
  author={Jiang, Changjiang and Yu, Fengchang and Chen, Haihua and Lu, Wei and Zeng, Jin},
  booktitle={Findings of the Association for Computational Linguistics: EMNLP 2025},
  pages={3172--3196},
  year={2025},
  doi={10.18653/v1/2025.findings-emnlp.169},
  ISBN={979-8-89176-335-7}
}

@inproceedings{fakehr1,
  author={Jiang, Changjiang and Sha, Xinkuan and Yu, Fengchang and Liu, Jingjing and Liu, Jian and Fang, Mingqi and Zhang, Chenfeng and Lu, Wei},
  booktitle={ICASSP 2026 - 2026 IEEE International Conference on Acoustics, Speech and Signal Processing (ICASSP)}, 
  title={Fake-HR1: Rethinking Reasoning of Vision Language Model for Synthetic Image Detection}, 
  year={2026},
  pages={10482--10486},
  doi={10.1109/ICASSP55912.2026.11462736}
}

@inproceedings{jiang2025ivy,
	author = {Jiang, Changjiang and Dong, Wenhui and Zhang, Zhonghao and Yu, Fengchang and Peng, Wei and Yuan, Xinbin and Bi, Yifei and Zhao, Ming and Zhou, Zian and Si, Chenyang and Shan, Caifeng},
	title = {Ivy-Fake: A Unified Explainable Framework and Benchmark for Image and Video AIGC Detection},
	year = {2026},
	isbn = {9798400726170},
	publisher = {Association for Computing Machinery},
	url = {https://doi.org/10.1145/3805622.3810615},
	doi = {10.1145/3805622.3810615},
	booktitle = {Proceedings of the 2026 International Conference on Multimedia Retrieval},
	pages = {2438–2447},
	numpages = {10},
	series = {ICMR '26}
}

@misc{meng2025mmeurekaexploringfrontiersmultimodal,
      title={MM-Eureka: Exploring the Frontiers of Multimodal Reasoning with Rule-based Reinforcement Learning}, 
      author={Fanqing Meng and Lingxiao Du and Zongkai Liu and Zhixiang Zhou and Quanfeng Lu and Daocheng Fu and Tiancheng Han and Botian Shi and Wenhai Wang and Junjun He and Kaipeng Zhang and Ping Luo and Yu Qiao and Qiaosheng Zhang and Wenqi Shao},
      year={2025},
      eprint={2503.07365},
      archivePrefix={arXiv},
      primaryClass={cs.CV},
      url={https://arxiv.org/abs/2503.07365}, 
}

@article{charxiv,
  title={Charxiv: Charting gaps in realistic chart understanding in multimodal llms},
  author={Wang, Zirui and Xia, Mengzhou and He, Luxi and Chen, Howard and Liu, Yitao and Zhu, Richard and Liang, Kaiqu and Wu, Xindi and Liu, Haotian and Malladi, Sadhika and others},
  journal={Advances in Neural Information Processing Systems},
  volume={37},
  pages={113569--113697},
  year={2024}
}

@article{zwz,
  title={Zooming without Zooming: Region-to-Image Distillation for Fine-Grained Multimodal Perception},
  author={Wei, Lai and He, Liangbo and Lan, Jun and Dong, Lingzhong and Cai, Yutong and Li, Siyuan and Zhu, Huijia and Wang, Weiqiang and Kong, Linghe and Wang, Yue and others},
  journal={arXiv preprint arXiv:2602.11858},
  year={2026}
}

@article{icot,
  title={Interleaved reasoning for large language models via reinforcement learning},
  author={Xie, Roy and Qiu, David and Gopinath, Deepak and Lin, Dong and Sun, Yanchao and Wang, Chong and Potdar, Saloni and Dhingra, Bhuwan},
  journal={arXiv preprint arXiv:2505.19640},
  year={2025}
}

@article{deepseekr1,
  title={Deepseek-r1: Incentivizing reasoning capability in llms via reinforcement learning},
  author={Guo, Daya and Yang, Dejian and Zhang, Haowei and Song, Junxiao and Wang, Peiyi and Zhu, Qihao and Xu, Runxin and Zhang, Ruoyu and Ma, Shirong and Bi, Xiao and others},
  journal={arXiv preprint arXiv:2501.12948},
  year={2025}
}

@inproceedings{fu-etal-2025-mitigating,
    title = "Mitigating Hallucination in Multimodal Large Language Model via Hallucination-targeted Direct Preference Optimization",
    author = "Fu, Yuhan  and
      Xie, Ruobing  and
      Sun, Xingwu  and
      Kang, Zhanhui  and
      Li, Xirong",
    editor = "Che, Wanxiang  and
      Nabende, Joyce  and
      Shutova, Ekaterina  and
      Pilehvar, Mohammad Taher",
    booktitle = "Findings of the Association for Computational Linguistics: ACL 2025",
    month = jul,
    year = "2025",
    address = "Vienna, Austria",
    publisher = "Association for Computational Linguistics",
    url = "https://aclanthology.org/2025.findings-acl.850/",
    doi = "10.18653/v1/2025.findings-acl.850",
    pages = "16563--16577",
    ISBN = "979-8-89176-256-5"
}

@misc{llava-onevision-1.5,
      title={LLaVA-OneVision-1.5: Fully Open Framework for Democratized Multimodal Training}, 
      author={Xiang An and Yin Xie and Kaicheng Yang and Wenkang Zhang and Xiuwei Zhao and Zheng Cheng and Yirui Wang and Songcen Xu and Changrui Chen and Chunsheng Wu and Huajie Tan and Chunyuan Li and Jing Yang and Jie Yu and Xiyao Wang and Bin Qin and Yumeng Wang and Zizhen Yan and Ziyong Feng and Ziwei Liu and Bo Li and Jiankang Deng},
      year={2025},
      eprint={2509.23661},
      archivePrefix={arXiv},
      primaryClass={cs.CV},
      url={https://arxiv.org/abs/2509.23661}, 
}

@inproceedings{
ChartMoE,
title={ChartMoE: Mixture of Diversely Aligned Expert Connector for Chart Understanding},
author={Zhengzhuo Xu and Bowen Qu and Yiyan Qi and SiNan Du and Chengjin Xu and Chun Yuan and Jian Guo},
booktitle={The Thirteenth International Conference on Learning Representations},
year={2025},
url={https://openreview.net/forum?id=o5TsWTUSeF}
}

@inproceedings{cppo,
  title={CPPO: Accelerating the Training of Group Relative Policy Optimization-Based Reasoning Models},
  author={Lin, ZhiHang and Lin, Mingbao and Xie, Yuan and Ji, Rongrong},
  booktitle={The Thirty-ninth Annual Conference on Neural Information Processing Systems},
  year={2025}
}

@inproceedings{codevision,
  title={Thinking with programming vision: Towards a unified view for thinking with images},
  author={Guo, Zirun and Hong, Minjie and Zhang, Feng and Jia, Kai and Jin, Tao},
  booktitle={Proceedings of the IEEE/CVF Conference on Computer Vision and Pattern Recognition},
  pages={33467--33476},
  year={2026}
}

@inproceedings{anls,
  title={Scene text visual question answering},
  author={Biten, Ali Furkan and Tito, Ruben and Mafla, Andres and Gomez, Lluis and Rusinol, Mar{\c{c}}al and Valveny, Ernest and Jawahar, CV and Karatzas, Dimosthenis},
  booktitle={Proceedings of the IEEE/CVF international conference on computer vision},
  pages={4291--4301},
  year={2019}
}

@article{grpo,
  title={Deepseekmath: Pushing the limits of mathematical reasoning in open language models},
  author={Shao, Zhihong and Wang, Peiyi and Zhu, Qihao and Xu, Runxin and Song, Junxiao and Bi, Xiao and Zhang, Haowei and Zhang, Mingchuan and Li, YK and Wu, Yang and others},
  journal={arXiv preprint arXiv:2402.03300},
  year={2024}
}

@article{dapo,
  title={Dapo: An open-source llm reinforcement learning system at scale},
  author={Yu, Qiying and Zhang, Zheng and Zhu, Ruofei and Yuan, Yufeng and Zuo, Xiaochen and Yue, Yu and Dai, Weinan and Fan, Tiantian and Liu, Gaohong and Liu, Lingjun and others},
  journal={arXiv preprint arXiv:2503.14476},
  year={2025}
}

@article{qi2026patchcue,
  title={PatchCue: Enhancing Vision-Language Model Reasoning with Patch-Based Visual Cues},
  author={Qi, Yukun and Fu, Pei and Li, Hang and Liu, Yuhan and Jiang, Chao and Qin, Bin and Luo, Zhenbo and Luan, Jian},
  journal={arXiv preprint arXiv:2603.05869},
  year={2026}
}

@misc{qwen36plus,
    title = {{Qwen3.6-Plus}: Towards Real World Agents},
    url = {https://qwen.ai/blog?id=qwen3.6},
    author = {{Qwen Team}},
    month = {April},
    year = {2026}
}

@misc{openai2024customjudge,
  title        = {Custom LLM as a Judge to Detect Hallucinations with Braintrust},
  author       = {{OpenAI}},
  year         = {2024},
  howpublished = {\url{https://developers.openai.com/cookbook/examples/custom-llm-as-a-judge/}},
  note         = {OpenAI Cookbook, accessed May 2026}
}

@misc{openrouter,
  title        = {OpenRouter Models},
  author       = {{OpenRouter}},
  year         = {2026},
  howpublished = {\url{https://openrouter.ai/models?input_modalities=image}},
  note         = {Accessed May 2026}
}

@inproceedings{veritas,
  title={Veritas: Generalizable Deepfake Detection via Pattern-Aware Reasoning},
  author={Tan, Hao and Lan, Jun and Tan, Zichang and Liu, Ajian and Song, Chuanbiao and Shi, Senyuan and Zhu, Huijia and Wang, Weiqiang and Wan, Jun and Lei, Zhen},
  booktitle={International Conference on Learning Representations},
  year={2026}
}

@article{veritasplus,
      title={Veritas++: Value-aware On-Policy Distillation for Perception-Enhanced AIGI Detection}, 
      author={Hao Tan and Jun Lan and Zichang Tan and Ajian Liu and Zijian Yu and Chuanbiao Song and Huijia Zhu and Weiqiang Wang and Jun Wan and Zhen Lei},
      year={2026},
      journal={arXiv preprint arXiv:2607.27113}
}

@inproceedings{earlvr,
    title = "Incentivizing Parametric Knowledge via Reinforcement Learning with Verifiable Rewards for Cross-Cultural Entity Translation",
    author = "Zhou, Jiang  and
      Zhao, Xiaohu  and
      Wu, Xinwei  and
      Dong, Tianyu  and
      Wang, Hao  and
      Liu, Yangyang  and
      Liu, Heng  and
      Xu, Linlong  and
      Wang, Longyue  and
      Luo, Weihua  and
      Xiong, Deyi",
    editor = "Liakata, Maria  and
      Moreira, Viviane P.  and
      Zhang, Jiajun  and
      Jurgens, David",
    booktitle = "Proceedings of the 64th Annual Meeting of the {A}ssociation for {C}omputational {L}inguistics (Volume 1: Long Papers)",
    month = jul,
    year = "2026",
    address = "San Diego, California, United States",
    publisher = "Association for Computational Linguistics",
    url = "https://aclanthology.org/2026.acl-long.254/",
    doi = "10.18653/v1/2026.acl-long.254",
    pages = "5616--5638",
    ISBN = "979-8-89176-390-6"
}

@inproceedings{cite_segui,
  author = {Yuan, Xinbin and Zhang, Jian and Li, Kaixin and Cai, Zhuoxuan and Yao, Lujian and Chen, Jie and Wang, Enguang and Hou, Qibin and Chen, Jinwei and Jiang, Peng-Tao and Li, Bo},
  booktitle = {Advances in Neural Information Processing Systems},
  pages = {127658--127679},
  publisher = {Curran Associates, Inc.},
  title = {SE-GUI: Enhancing Visual Grounding for GUI Agents via Self-Evolutionary Reinforcement Learning},
  url = {https://proceedings.neurips.cc/paper_files/paper/2025/file/b95c7e24501f5d1dddbc5e8526cda7ae-Paper-Conference.pdf},
  volume = {38},
  year = {2025},
}

@misc{cite_fakevlmr1,
  title = {FakeVLM-R1: Internalizing Physical Laws via CoT for Synthetic Image Detection},
  author = {Leqi Zhu and Junyan Ye and Kaiqing Lin and Zhiyuan Yan and Conghui He and Weijia Li},
  year = {2026},
  eprint = {2605.30062},
  archiveprefix = {arXiv},
  primaryclass = {cs.CV},
  url = {https://arxiv.org/abs/2605.30062},
}

@article{zhao2026know,
  title={Know What You Know: Metacognitive Entropy Calibration for Verifiable RL Reasoning},
  author={Zhao, Qiannian and Yang, Chen and Jing, Jinhao and Zhang, Yunke and Ren, Xuhui and Yu, Lu and Zhang, Shijie and Yin, Hongzhi},
  journal={arXiv preprint arXiv:2602.22751},
  year={2026}
}

@inproceedings{qu2023towards,
  title={Towards robust tampered text detection in document image: New dataset and new solution},
  author={Qu, Chenfan and Liu, Chongyu and Liu, Yuliang and Chen, Xinhong and Peng, Dezhi and Guo, Fengjun and Jin, Lianwen},
  booktitle={2023 IEEE/CVF Conference on Computer Vision and Pattern Recognition (CVPR)},
  pages={5937--5946},
  year={2023},
  organization={IEEE}
}

@inproceedings{qu2024towards,
  title={Towards modern image manipulation localization: A large-scale dataset and novel methods},
  author={Qu, Chenfan and Zhong, Yiwu and Liu, Chongyu and Xu, Guitao and Peng, Dezhi and Guo, Fengjun and Jin, Lianwen},
  booktitle={2024 IEEE/CVF Conference on Computer Vision and Pattern Recognition (CVPR)},
  pages={10781--10790},
  year={2024},
  organization={IEEE}
}

@inproceedings{qu2025revisiting,
  title={Revisiting tampered scene text detection in the era of generative AI},
  author={Qu, Chenfan and Zhong, Yiwu and Guo, Fengjun and Jin, Lianwen},
  booktitle={Proceedings of the AAAI Conference on Artificial Intelligence},
  volume={39},
  number={1},
  pages={694--702},
  year={2025}
}

@inproceedings{qu2026omni,
  title={Omni-IML: Towards Unified Interpretable Image Manipulation Localization},
  author={Qu, Chenfan and Zhong, Yiwu and Guo, Fengjun and Jin, Lianwen},
  booktitle={The Fourteenth International Conference on Learning Representations},
  year={2026}
}

@inproceedings{qu2026detect,
  title={Detect Any AI-Counterfeited Text Image},
  author={Qu, Chenfan and Zhong, Yiwu and Zhu, Xuekang and Li, Junchi and Jiang, Changjiang and Jin, Lianwen and others},
  booktitle={Proceedings of the IEEE/CVF Conference on Computer Vision and Pattern Recognition},
  pages={35437--35450},
  year={2026}
}

@inproceedings{zhu2025mesoscopic,
  title={Mesoscopic insights: orchestrating multi-scale \& hybrid architecture for image manipulation localization},
  author={Zhu, Xuekang and Ma, Xiaochen and Su, Lei and Jiang, Zhuohang and Du, Bo and Wang, Xiwen and Lei, Zeyu and Feng, Wentao and Pun, Chi-Man and Zhou, Ji-Zhe},
  booktitle={Proceedings of the AAAI conference on artificial intelligence},
  volume={39},
  number={10},
  pages={11022--11030},
  year={2025}
}

@article{ma2024imdl,
  title={Imdl-benco: A comprehensive benchmark and codebase for image manipulation detection \& localization},
  author={Ma, Xiaochen and Zhu, Xuekang and Su, Lei and Du, Bo and Jiang, Zhuohang and Tong, Bingkui and Lei, Zeyu and Yang, Xinyu and Pun, Chi-Man and Lv, Jiancheng and others},
  journal={Advances in Neural Information Processing Systems},
  volume={37},
  pages={134591--134613},
  year={2024}
}

@article{du2026forensichub,
  title={Forensichub: A unified benchmark \& codebase for all-domain fake image detection and localization},
  author={Du, Bo and Zhu, Xuekang and Ma, Xiaochen and Qu, Chenfan and Feng, Kaiwen and Yang, Zhe and Pun, Chi-Man and Zhou, Ji-Zhe and others},
  journal={Advances in neural information processing systems},
  volume={38},
  year={2026}
}

@inproceedings{zhu2026revisiting,
  title={Revisiting image manipulation localization under realistic manipulation scenarios},
  author={Zhu, Xuekang and Zhou, Ji-Zhe and Feng, Kaiwen and Qu, Chenfan and Wang, Xiwen and Wang, Yunfei and Zhou, Liting and Liu, Jian},
  booktitle={Proceedings of the IEEE/CVF Conference on Computer Vision and Pattern Recognition},
  pages={7198--7207},
  year={2026}
}

@article{DeFakerOne,
  title = {Venus-DeFakerOne: Unified Fake Image Detection \& Localization},
  author = {Team, GuangJian},
  journal = {arXiv preprint arXiv:2605.14091},
  year = {2026},
}

@article{ying2026beyond,
  title={Beyond Human Annotation: Recent Advances in Data Generation Methods for Document Intelligence},
  author={Ying, Dehao and Yu, Fengchang and Chen, Haihua and Jiang, Changjiang and Li, Yurong and Lu, Wei},
  journal={arXiv preprint arXiv:2601.12318},
  year={2026}
}

@article{xin2026unimomo,
  title={UniMoMo: Expert Merging-Based MoE Acceleration for Large Recommendation Models},
  author={Xin, Lei and Gu, Bin and Li, Peize and Wang, Zitong and Zhao, Jianbo and Jiang, Changjiang and Xie, Yanyue and Huang, Chao and Zhao, Xuyang and Su, Zunhai and others},
  journal={arXiv preprint arXiv:2608.08627},
  year={2026}
}

@article{xin2026hytrec,
  title={Hytrec: A hybrid temporal-aware attention architecture for long behavior sequential recommendation},
  author={Xin, Lei and Zheng, Yuhao and Cheng, Ke and Jiang, Changjiang and Zhang, Zifan and Zeng, Fanhu},
  journal={arXiv preprint arXiv:2602.18283},
  year={2026}
}

@inproceedings{lin2025audio,
  title={Audio does matter: Importance-aware multi-granularity fusion for video moment retrieval},
  author={Lin, Junan and Liu, Daizong and Chen, Xianke and Qu, Xiaoye and Yang, Xun and Zhu, Jixiang and Zhang, Sanyuan and Dong, Jianfeng},
  booktitle={Proceedings of the 33rd ACM International Conference on Multimedia},
  pages={6027--6036},
  year={2025}
}

@misc{xin2026dualcpt,
  title={DualCPT: Dual-branch Modeling for Cellular Phenotype Transition},
  author={Xin, Lei and Kong, Zhenglun and Chen, Fukang and Zheng, Yuhao and Wang, Zeheng and Tang, Hao},
  journal={AAAI Bridge Program on AI for Medicine and Healthcare},
  pages={302--312},
  year={2026},
  publisher={PMLR}
}

@article{xin2024artificial,
  title={Artificial intelligence for central dogma-centric multi-omics: Challenges and breakthroughs},
  author={Xin, Lei and Huang, Caiyun and Li, Hao and Huang, Shihong and Feng, Yuling and Kong, Zhenglun and Liu, Zicheng and Li, Siyuan and Yu, Chang and Shen, Fei and others},
  journal={arXiv preprint arXiv:2412.12668},
  year={2024}
}

@article{kong2025token,
  title={Token Reduction Should Go Beyond Efficiency in Generative Models--From Vision, Language to Multimodality},
  author={Kong, Zhenglun and Li, Yize and Zeng, Fanhu and Xin, Lei and Messica, Shvat and Lin, Xue and Zhao, Pu and Kellis, Manolis and Tang, Hao and Zitnik, Marinka},
  journal={arXiv preprint arXiv:2505.18227},
  year={2025}
}
